\documentclass[DIV15, a4paper, 11pt]{article}

\usepackage[ruled, vlined, linesnumbered, commentsnumbered]{algorithm2e}
\usepackage[toc,page]{appendix}
\usepackage{authblk}
\usepackage{prettyref}
\usepackage{subfig}
\usepackage{graphicx}
\usepackage{booktabs}
\usepackage{colortbl}
\usepackage{amsthm,amsopn,amssymb,amsmath,amsfonts}
\usepackage{algorithmic}
\usepackage{multirow}
\usepackage{tabularx}
\usepackage{footnote}
\usepackage{threeparttable}
\usepackage{hhline}
\usepackage{cite}
\usepackage{scrpage2}
\usepackage{setspace}
\usepackage{color}
\usepackage{xcolor}  

\usepackage[margin=2.2cm]{geometry}	

\usepackage{tikz,pgfplots}

\definecolor{myblue}{RGB}{34,31,217}
\definecolor{mycyan}{gray}{.7}

\newtheorem{definition}{Definition}

\DeclareMathOperator*{\argmin}{argmin}

\newcommand{\pref}{\prettyref}

\newrefformat{fig}{Fig.~\ref{#1}}
\newrefformat{tab}{Table~\ref{#1}}
\newrefformat{sec}{Section~\ref{#1}}
\newrefformat{alg}{Algorithm~\ref{#1}}
\newrefformat{property}{Property~\ref{#1}}
\newrefformat{theorem}{Theorem~\ref{#1}}
\newrefformat{def}{Definition~\ref{#1}}
\newrefformat{eq}{equation~(\ref{#1})}
\newrefformat{app}{Appendix~\ref{#1}}
\newrefformat{proposition}{Proposition~\ref{#1}}

\usepackage{lscape}

\begin{document}

\title{\textbf\LARGE\selectfont Evolutionary Many-Objective Optimization Based on Adversarial Decomposition}

\author[1]{\normalsize\fontfamily{lmss}\selectfont Mengyuan Wu}
\author[2]{\normalsize\fontfamily{lmss}\selectfont Ke Li}
\author[1]{\normalsize\fontfamily{lmss}\selectfont Sam Kwong}
\author[1]{\normalsize\fontfamily{lmss}\selectfont Qingfu Zhang}
\affil[1]{\normalsize\fontfamily{lmss}\selectfont Department of Computer Science, City University of Hong Kong}
\affil[2]{\normalsize\fontfamily{lmss}\selectfont College of Engineering, Mathematics and Physical Sciences, University of Exeter}
\affil[*]{\normalsize\fontfamily{lmss}\selectfont Email: k.li@exeter.ac.uk, mengyuan.wu@my.cityu.edu.hk, {cssamk, qingfu.zhang}@cityu.edu.hk}

\date{}
\maketitle

{\normalsize\fontfamily{lmss}\selectfont \textbf{Abstract:}}
The decomposition-based method has been recognized as a major approach for multi-objective optimization. It decomposes a multi-objective optimization problem into several single-objective optimization subproblems, each of which is usually defined as a scalarizing function using a weight vector. Due to the characteristics of the contour line of a particular scalarizing function, the performance of the decomposition-based method strongly depends on the Pareto front's shape by merely using a single scalarizing function, especially when facing a large number of objectives. To improve the flexibility of the decomposition-based method, this paper develops an adversarial decomposition method that leverages the complementary characteristics of two different scalarizing functions within a single paradigm. More specifically, we maintain two co-evolving populations simultaneously by using different scalarizing functions. In order to avoid allocating redundant computational resources to the same region of the Pareto front, we stably match these two co-evolving populations into one-one solution pairs according to their working regions of the Pareto front. Then, each solution pair can at most contribute one mating parent during the mating selection process. Comparing with nine state-of-the-art many-objective optimizers, we have witnessed the competitive performance of our proposed algorithm on 130 many-objective test instances with various characteristics and Pareto front's shapes.

{\normalsize\fontfamily{lmss}\selectfont \textbf{Keywords:}} Many-objective optimization, evolutionary algorithm, stable matching theory, decomposition-based method.


\section{Introduction}
\label{sec:introduction}

Many real-life disciplines (e.g., optimal design~\cite{RopponenRP11}, economics~\cite{EconomicsApp} and water management~\cite{ReedH14}) often involve optimization problems having multiple conflicting objectives, known as multi-objective optimization problems (MOPs). Rather than a global optimum that optimizes all objectives simultaneously, in multi-objective optimization, decision makers often look for a set of Pareto-optimal solutions which consist of the best trade-offs among conflicting objectives. The balance between convergence and diversity is the cornerstone of multi-objective optimization. In particular, the convergence means the approximation to the Pareto-optimal set should be as close as possible; while the diversity means the spread of the trade-off solutions should be as uniform as possible.

Evolutionary algorithm, which in principle can approximate the Pareto-optimal set in a single run, has been widely accepted as a major approach for multi-objective optimization~\cite{DebBook}. Over the past three decades, many efforts have been devoted in developing evolutionary multi-objective optimization (EMO) algorithms and have obtained recognized performance on problems with two or three objectives~\cite{DebAPM02,SPEA2,SMSEMOA,SMC12,INS12,INS13,NEUCOM14,LiFKZ14}. With the development of science and technology, real-life optimization scenarios bring more significant challenges, e.g., complicated landscape, multi-modality and high dimensionality, to the algorithm design. As reported in \cite{WagnerBN06,IshibuchiTN08,LiLTY15}, the performance of traditional EMO algorithms severely deteriorate with the increase of the number of objectives. Generally speaking, researchers owe the performance deterioration to three major reasons, i.e., the loss of selection pressure for Pareto domination~\cite{IshibuchiTN08}, the difficulty of density estimation in a high-dimensional space~\cite{DebJ14} and its anti-convergence phenomenon~\cite{AdraF11}, and the exponentially increased computational complexity~\cite{BaderZ11}.

As a remedy to the loss of selection pressure for Pareto domination in a high-dimensional space, many modified dominance relationships have been developed to strengthen the comparability between solutions, e.g., $\epsilon$-dominance~\cite{BORG}, fuzzy Pareto-dominance~\cite{HeYZ14}, $k$-optimality~\cite{FarinaA04}, preference order ranking~\cite{PierroKS07}, control of dominance area~\cite{SatoAT06} and grid dominance~\cite{GrEA}. Very recently, a generalized Pareto-optimality was developed in~\cite{ZhuXG16} to expand the dominance area of Pareto domination, so that the percentage of non-dominated solutions in a population does not increase dramatically with the number of objectives. Different from~\cite{SatoAT06}, the expansion envelop for all solutions are kept the same in~\cite{ZhuXG16}. Other than the Pareto dominance-based approaches, $L$-optimality was proposed in~\cite{ZouCLK08} to help rank the solutions.

The loss of selection pressure can also be remedied by an effective diversity maintenance strategy. To relieve the anti-convergence phenomenon, \cite{AdraF11} applied the maximum spread indicator developed in \cite{Zitzler99} to activate and deactivate the diversity promotion mechanism in NSGA-II~\cite{DebAPM02}. To facilitate the density estimation in a high-dimensional space, \cite{DebJ14} proposed to replace the crowding distance used in NSGA-II by counting the number of associated solutions with regard to a predefined reference point. In particular, a solution is considered being associated with a reference point if it has the shortest perpendicular distance to this reference point. Instead of handling the convergence and the diversity separately, \cite{WangPF13} proposed to co-evolve a set of target vectors, each of which represents a optimization goal. The fitness value of a solution is defined by aggregating the number of current goals it achieves, i.e., dominates. The population diversity is implicitly maintained by the goals widely spread in the objective space; while the selection pressure towards the convergence is strengthened by co-evolving the optimization goals with the population.

The exponentially increased computational costs come from two aspects: 1) the exponentially increased computational complexity for calculating the hypervolume, which significantly hinders the scalability of the indicator-based EMO algorithms; 2) the significantly increased computational costs for maintaining the non-domination levels~\cite{DebAPM02} of a large number of solutions in a high-dimensional space. As for the former aspect, some improved methods, from the perspective of computational geometry~\cite{BeumeFLPV09,BringmannF10,WhileBB12} or Monte carlo sampling~\cite{BaderZ11}, have been proposed to enhance the efficiency of hypervolume calculation. As for the latter aspect, many efforts have been devoted to applying some advanced data structures to improve the efficiency of non-dominated sorting procedure~\cite{ZhangTCJ15,ZhouCZ17,GustavssonS17}. It is worth noting that our recent study~\cite{LiDZZ16} showed that it can be more efficient to update the non-domination levels by leveraging the population structure than to sort the population from scratch in every iteration.

As reported in~\cite{IshibuchiSTN09,IshibuchiAN15,IshibuchiSMN16}, decomposition-based EMO methods have become increasingly popular for solving problems with more than three objectives, which are often referred to as many-objective optimization problems (MaOPs). Since the decomposition-based EMO methods transform an MOP into several single-objective optimization subproblems, it does not suffer the loss of selection pressure of Pareto domination in a high-dimensional space. In addition, the update of the population relies on the comparison of the fitness values, thus the computational costs do not excessively increase with the dimensionality. As reported in \cite{ZhouZ16}, different subproblems, which focus on different regions in the objective space, have various difficulties. Some superior solutions can easily dominantly occupy several subproblems. This is obviously harmful to the population diversity and getting worse with the increase of the dimensionality. To overcome this issue, \cite{LiKZD15} built an interrelationship between subproblems and solutions, where each subproblem can only be updated by its related solutions. Based on the similar merit, \cite{MOEADDU} restricted a solution to only updating one of its $K$ closest subproblems. In \cite{AsafuddoulaRS15}, two metrics were proposed to measure the convergence and diversity separately. More specifically, the objective vector of a solution is at first projected onto its closest weight vector. Then, the distance between the projected point and the ideal point measures the solution's convergence; while the distance between the projected point and the original objective vector measures the solution's diversity. At the end, a diversity-first update scheme was developed according to these two metrics. Analogously, \cite{HeY16} developed a modified achievement scalarizing function as the convergence metric while an angle-based density estimation method was employed to measure the solution's diversity.

Recently, there is a growing trend in leveraging the advantages of the decomposition- and Pareto-based methods within the same framework. For example, \cite{LiDZK15} suggested to use the Pareto domination to prescreen the population. The local density is estimated by the number of solutions associated with a pre-defined weight vector. In particular, a solution located in an isolated region has a higher chance to survive to the next iteration. Differently, \cite{JiangY16} developed an angle-based method to estimate the local crowdedness around a weight vector. In addition to the density estimation, the weight vectors divide the objective space into different subspaces, which is finally used to estimate the local strength value~\cite{SPEA2} of each solution. Analogously, in \cite{YuanXWY16}, a non-dominated sorting is conducted upon all the subspaces, where solutions in different subspaces are considered non-dominated to each other. \cite{LiKD15} developed a dual-population paradigm which co-evolves two populations simultaneously. These populations are maintained by different selection mechanisms respectively, while their interaction is implemented by a restricted mating selection mechanism.

Although the decomposition-based EMO methods have been extensively used for MaOPs, a recent comprehensive study~\cite{IshibuchiSMN16} demonstrated that the performance of decomposition-based EMO methods strongly depends on the shape of the Pareto front (PF). This phenomenon can be attributed to two major reasons:
\begin{itemize}
    \item Most, if not all, decomposition-based EMO methods merely consider a single scalarizing function in fitness assignment. Since the contour line of a scalarizing function does not adapt to a particular problem's characteristic, the flexibility is restricted.
    \item As discussed in the previous paragraph, different regions of the PF have various difficulties. The balance between convergence and diversity of the search process can be easily biased by some dominantly superior solutions. The increasing dimensionality exacerbates this phenomenon.
\end{itemize}
Bearing the above considerations in mind, this paper develops a new decomposition-based method, called adversarial decomposition, for many-objective optimization. Generally speaking, it has the following three features:
\begin{itemize}
    \item It maintains two co-evolving populations simultaneously, each of which is maintained by a different scalarizing function. In particular, one population uses the boundary intersection-based scalarizing function, while the other one applies a modified achievement scalarizing function. In this regard, the search behaviors of these two populations become different, where one is convergence-oriented while the other is diversity-oriented.
	\item In order to make these two populations complement each other, they use ideal and nadir points respectively as the reference point in their scalarizing functions. By doing this, the two populations are evolved following two sets of adversarial search directions.
	\item During the mating selection process, two populations are at first stably matched to form a set of one-one solution pairs. In particular, solutions within the same pair concentrate on similar regions of the PF. Thereafter, each solution pair can at most contribute one mating parent for offspring generation. By doing this, we can expect an uniformly spread of the computational efforts over the entire PF.
\end{itemize} 

The remainder of this paper is organized as follows. \pref{sec:preliminary} provides some preliminaries useful to this paper. \pref{sec:proposal} describes the technical details of our proposed method step by step. The empirical studies are presented and discussed in~\pref{sec:setup} and~\pref{sec:experiments}. Finally, \pref{sec:conclusion} concludes the paper and provides some future directions.


\section{Preliminaries}
\label{sec:preliminary}

\begin{figure*}[!t]
	\centering
	\subfloat[TCH]{\includegraphics[width=.33\linewidth]{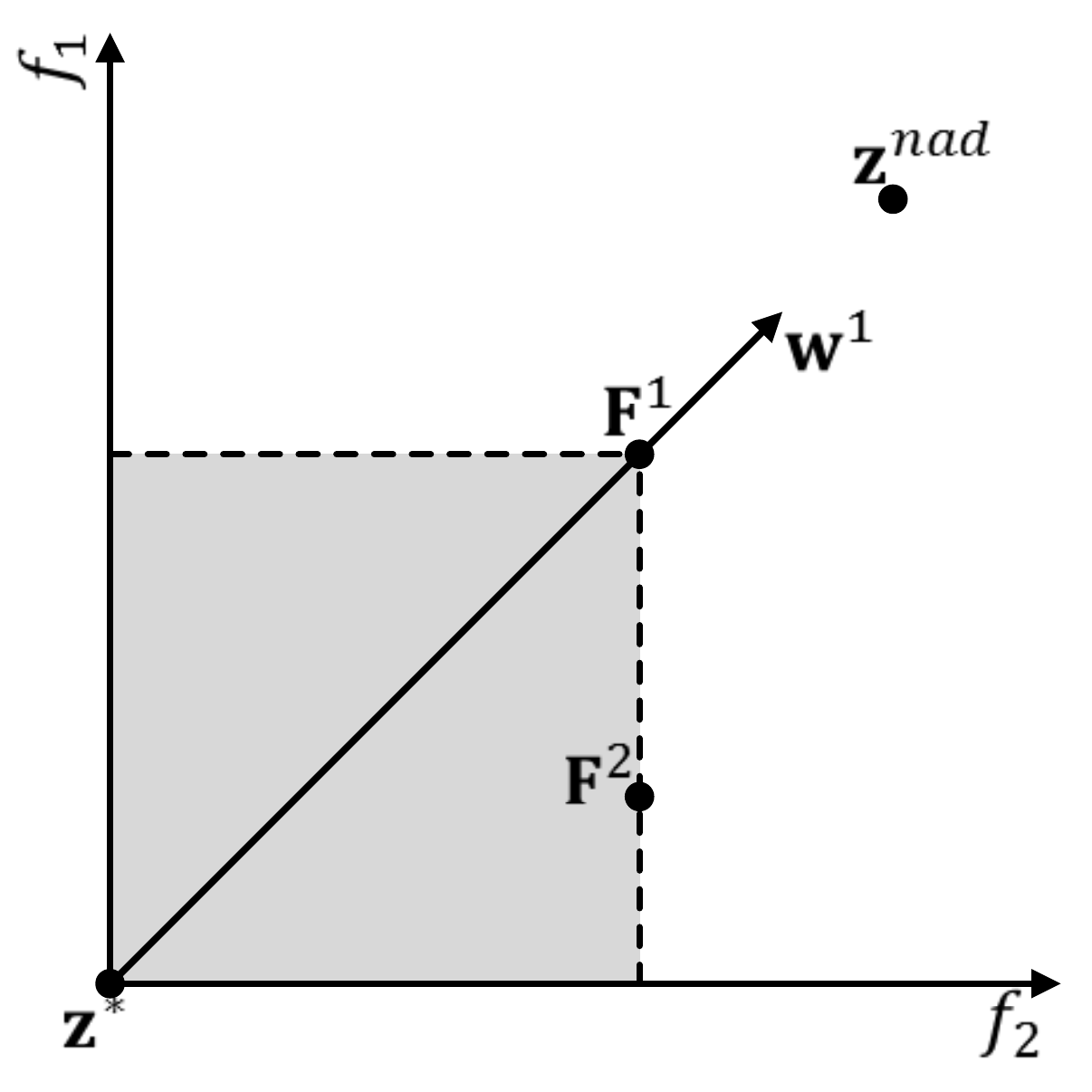}}
	\subfloat[PBI]{\includegraphics[width=.33\linewidth]{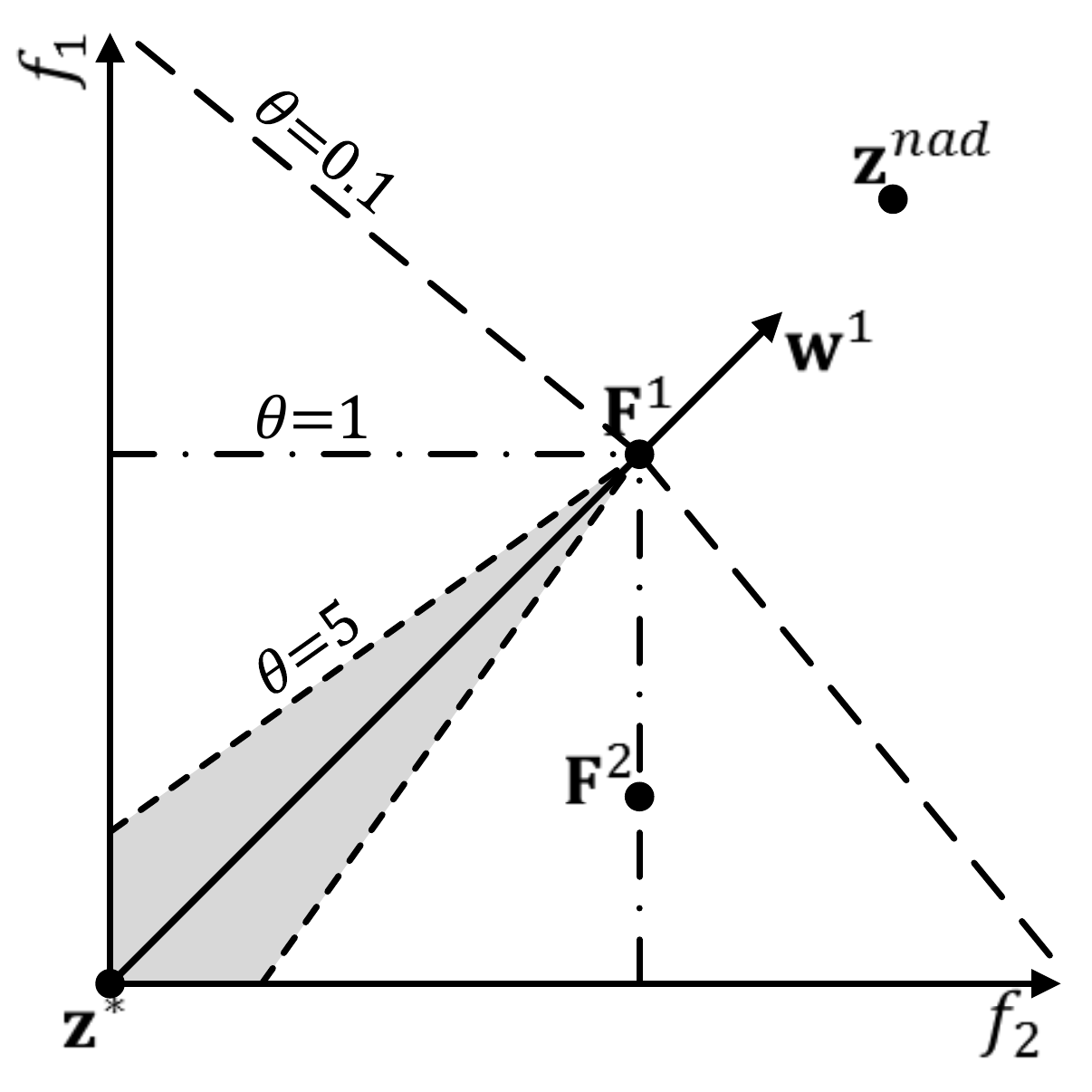}}
	\subfloat[MA-ASF]{\includegraphics[width=.33\linewidth]{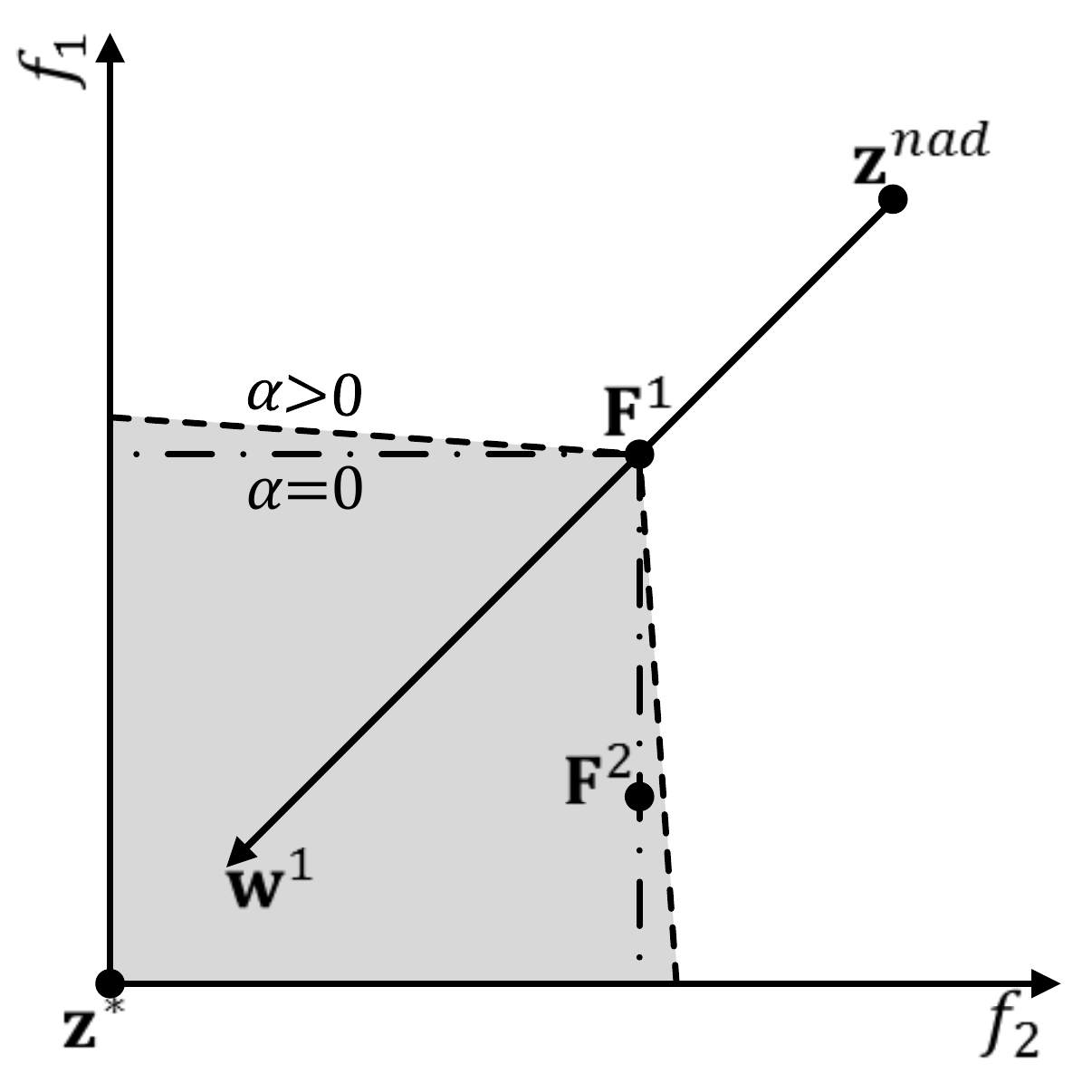}}
	\caption{Illustration of the characteristics of different scalarizing functions.}
	\label{fig:decomposition}
\end{figure*}

This section first provides some basic definitions of multi-objective optimization. Afterwards, we briefly introduce the decomposition of multi-objective optimization.

\subsection{Basic Definitions}
\label{sec:MO}

The MOP considered in this paper can be mathematically defined as follows:
\begin{equation}
\begin{array}{l}
\mathrm{minimize} \quad \mathbf{F}(\mathbf{x})=(f_1(\mathbf{x}),\cdots,f_m(\mathbf{x}))^{T}\\
\mathrm{subject\ to} \quad \mathbf{x}\in\Omega
\end{array},
\label{eq:MOP}
\end{equation}
where $\mathbf{x}=(x_1,\cdots,x_n)^T\in\Omega$ is a $n$-dimensional decision variable vector from the decision space $\mathbb{R}^n$ and $\mathbf{F}(\mathbf{x})$ is a $m$-dimensional objective vector in the objective space $\mathbb{R}^m$. 

\begin{definition}
Given solutions $\mathbf{x}^1,\mathbf{x}^2\in\Omega$, $\mathbf{x}^1$ is said to \textit{dominate} $\mathbf{x}^2$, denoted by $\mathbf{x}^1\preceq\mathbf{x}^2$, if and only if $f_i(\mathbf{x}^1)\leq f_i(\mathbf{x}^2)$ for all $i\in\{1,\cdots,m\}$ and $\mathbf{F}(\mathbf{x}^1)\neq \mathbf{F}(\mathbf{x}^2)$.
\label{def:pareto}
\end{definition}

\begin{definition}
A solution $\mathbf{x}\in\Omega$ is called \textit{Pareto-optimal} if and only if there is no other solution dominates it.
\end{definition}

\begin{definition}
The \textit{Pareto-optimal set} (PS) is defined as the set of all Pareto-optimal solutions. Their corresponding objective vectors form the \textit{Pareto-optimal front} (PF).
\end{definition}

\begin{definition}
The \textit{ideal point} is $\mathbf{z}^\ast=(z_1^\ast,\cdots,z_m^\ast)^T$, where $z_i^\ast=\min\limits_{\mathbf{x}\in\Omega}\{f_i(\mathbf{x})\}$ for all $i\in\{1,\cdots,m\}$. The \textit{nadir point} is $\mathbf{z}^{nad}=(z_1^{nad},\cdots,z_m^{nad})^T$, where $z_i^{nad}=\max\limits_{\mathbf{x}\in\Omega}\{f_i(\mathbf{x})\}$.
\label{def:ideal}
\end{definition}

\subsection{Decomposition}
\label{sec:decomposition}

Under some mild conditions, the task of approximating the PF can be decomposed into several scalar optimization subproblems, each of which is formed by a weighted aggregation of all individual objectives~\cite{MOEAD}. In the classic multi-objective optimization literature~\cite{nonlinear}, there have been several established approaches for constructing scalarizing functions, among which the weighted Tchebycheff (TCH) and penalty-based boundary intersection (PBI)~\cite{MOEAD} are the most popular ones. More specifically, the TCH function is mathematically defined as:
\begin{equation}
\begin{array}{l l}
\mathrm{minimize}\quad g^{tch}(\mathbf{x}|\mathbf{w},\mathbf{z}^{\ast})=\max\limits_{1\leq i\leq m}\{|f_i(\mathbf{x})-z_{i}^{\ast}|/w_i\}\\
\mathrm{subject\ to}\quad \mathbf{x}\in\Omega
\end{array},
\label{eq:TCH}
\end{equation}
where $\mathbf{w}=(w^1,\cdots,w^m)^T$ is a user specified weight vector, $w_i\geq 0$ for all $i\in\{1,\cdots,m\}$ and $\sum_{i=1}^m w_i=1$. Note that $w_i$ is set to be a very small number, say $10^{-6}$, in case $w_i=0$. The contour line of the TCH function is shown in~\pref{fig:decomposition}(a) where $\mathbf{w}=(0.5,0.5)^T$. From this figure, we can clearly see that the control area (i.e., the area that holds better solutions) of the TCH function is similar to the Pareto domination defined in~\pref{def:pareto}, e.g., solutions located in the gray shaded area (i.e., the control area of $\mathbf{F}^1$) are better than $\mathbf{F}^1$. Note that the TCH function cannot discriminate the \textit{weakly} dominated solution~\cite{nonlinear}. For example, the TCH function values of $\mathbf{F}^1$ and $\mathbf{F}^2$ are the same, but $\mathbf{F}^1\preceq\mathbf{F}^2$ according to \pref{def:pareto}.

As for the PBI function, it is mathematically defined as:
\begin{equation}
\begin{array}{l}
\begin{split}
\mathrm{minimize}\quad g^{pbi}(\mathbf{x}|\mathbf{w},\mathbf{z}^{\ast})&=d_1(\mathbf{F}(\mathbf{x})|\mathbf{w},\mathbf{z}^\ast)\\
&+\theta d_2(\mathbf{F}(\mathbf{x})|\mathbf{w},\mathbf{z}^\ast)\\
\end{split}\\
\mathrm{subject\ to}\quad \mathbf{x}\in\Omega
\end{array},
\label{eq:pbi}
\end{equation}
where
\begin{equation}
\begin{array}{l}
d_1(\mathbf{y}|\mathbf{w},\mathbf{z})=\dfrac{\|(\mathbf{y}-\mathbf{z})^T\mathbf{w}\|}{\|\mathbf{w}\|}\\
d_2(\mathbf{y}|\mathbf{w},\mathbf{z})=\|\mathbf{y}-(\mathbf{z}+\dfrac{d_1}{\|\mathbf{w}\|}\mathbf{w})\|\\
\end{array}.
\label{eq:d1}
\end{equation}
As discussed in~\cite{LiDZK15}, $d_1$ and $d_2$ measure the convergence and diversity of $\mathbf{x}$ with regard to $\mathbf{w}$, respectively. The balance between convergence and diversity is parameterized by $\theta$, which also controls the contour line of the PBI function. In~\pref{fig:decomposition}(b), we present the contour lines of PBI functions with different $\theta$ settings.


\section{Many-Objective Optimization Algorithm Based on Adversarial Decomposition}
\label{sec:proposal}

In this section, we introduce the technical details of our proposed many-objective optimization algorithm based on adversarial decomposition, denoted as MOEA/AD whose pseudo code is given in~\pref{alg:moead-2p}, step by step. At the beginning, we initialize a population of solutions $S=\{\mathbf{x}^1,\cdots,\mathbf{x}^N\}$ via random sampling upon $\Omega$; the ideal and nadir points; a set of weight vectors $W=\{\mathbf{w}^1,\cdots,\mathbf{w}^N\}$ and build their neighborhood structure according to the method in~\cite{LiDZK15}. Afterwards, we assign $S$ directly to the two co-evolving populations, i.e., diversity population $S_d$ and convergence population $S_c$. Note that $S_d$ and $S_c$ share the same weight vectors, each of which corresponds to a unique subproblem for $S_d$ and $S_c$ respectively. To facilitate the mating selection process, we initialize a matching array $M$ and a sentinel array $R$, where $M[i]$ indicates a solution $\mathbf{x}_d^i$ in $S_d$ is paired with a solution $\mathbf{x}_c^{M[i]}$ in $S_c$ and $R[i]$ indicates whether this pair of solutions work in similar regions of the PF. During the main while loop, the mating parents are selected from the solution pairs. The generated offspring solution is used to update $S_d$ and $S_c$ separately. After each generation, we divide the solutions in $S_d\cup S_c$ into different solution pairs for the next round's mating selection process. The major components of MOEA/AD are explained in the following subsections.

\begin{algorithm}[!t]
	\caption{$\mathsf{MOEA/AD}$}
	\label{alg:moead-2p}
	\KwIn{algorithm parameters}
	\KwOut{final population $S_c$ and $S_d$}
	Initialize a set of solutions $S$, $\mathbf{z}^\ast$ and $\mathbf{z}^{nad}$;\\
	Initialize a set of weight vectors $W$ and its neighborhood structure $B$;\\
	$S_c\leftarrow S$, $S_d\leftarrow S$;\\
	\For{$i\leftarrow 1$ \KwTo $N$}{
		$M[i]\leftarrow i$, $R[i]\leftarrow 1$;\\
	}
	$generation\leftarrow 0$;\\
	\While{Stopping criterion is not satisfied}{
		\For{$i\leftarrow 1$ \KwTo $N$}{
			$\overline{S}\leftarrow\mathsf{MatingSelection}(S_c,S_d,i,M,R,W,B)$;\\
			$\overline{\mathbf{x}}\leftarrow\mathsf{Variation}(\overline{S})$;\\
			Update $\mathbf{z}^\ast$ and $\mathbf{z}^{nad}$;\\
			$(S_c,S_d)\leftarrow\mathsf{PopulationUpdate}(S_c,S_d,\overline{\mathbf{x}},W)$;\\
		}
		$(M,R)\leftarrow\mathsf{Match}(S_c,W)$;\\
		$generation$++;\\
	}
	\Return $S_c$, $S_d$;
\end{algorithm}

\subsection{Adversarial Decomposition}
\label{sec:ad}

As discussed in~\pref{sec:introduction}, the flexibility of the decomposition-based method is restricted due to the use of a single scalarizing function. Bearing this consideration in mind, this paper develops an adversarial decomposition method. Its basic idea is to maintain two co-evolving and complementary populations simultaneously, each of which is maintained by a different scalarizing function.

More specifically, one population is maintained by the PBI function introduced in~\pref{sec:decomposition}, where we set $\theta=5.0$ as recommended in~\cite{LiDZK15}. The other population is maintained by a modified augmented achievement scalarizing function (MA-ASF) defined as follows:
\begin{equation}
\begin{array}{l}
\begin{split}
\mathrm{minimize}\quad g^{c}(\mathbf{x}|\mathbf{w},\mathbf{z}^{nad})&=\max\limits_{1\leq i\leq m}\{(f_i(\mathbf{x})-z_{i}^{nad})/{w_i}\}\\
&+\alpha\sum\limits_{i=1}^{m}{(f_i(\mathbf{x})-z_{i}^{nad})/{w_i}}\\
\end{split}\\
\mathrm{subject\ to}\quad \mathbf{x}\in\Omega
\end{array},
\label{eq:maasf}
\end{equation}
where $\alpha$ is an augmentation coefficient. Comparing with the TCH function in~\pref{eq:TCH}, the MA-ASF uses the nadir point to replace the ideal point and the absolute operator is removed to allow $f_i(\mathbf{x})$ to be smaller than $z_{i}^{nad}$, where $i\in\{1,\cdots,m\}$. Furthermore, the augmentation term in the MA-ASF helps avoid weakly Pareto-optimal solutions. As shown in~\pref{fig:decomposition}(c), the contour line of the MA-ASF is the same as that of the TCH function in case $\alpha=0$; while the control area of the MA-ASF becomes wider when setting $\alpha>0$. In this case, the MA-ASF is able to discriminate the weakly dominated solution, e.g., the MA-ASF value of $\mathbf{F}^2$ in \pref{fig:decomposition}(c) is better than that of $\mathbf{F}^1$ when $\alpha>0$. Here we use a sufficiently small $\alpha=10^{-6}$ as recommended in~\cite{ASF}.

To deal with problems having different scales of objectives, we normalize the objective values before using the scalarizing function. By doing this, the PBI function becomes:
\begin{equation}
    \overline{g}^{d}(\mathbf{x}|\mathbf{w})=d_1(\overline{\mathbf{F}}(\mathbf{x})|\mathbf{w},\textbf{0})+\theta d_2(\overline{\mathbf{F}}(\mathbf{x})|\mathbf{w},\textbf{0}),
\label{eq:npbi}
\end{equation}
where $\overline{\mathbf{F}}(\mathbf{x})=(\overline{f_1}(\mathbf{x}),\cdots,\overline{f_m}(\mathbf{x}))^T$ and $\overline{f_i}(\mathbf{x})=\frac{f_i(\mathbf{x})-z^{\ast}_i}{z^{nad}_i-z^{\ast}_i}$ for all $i\in\{1,\cdots,m\}$. The MA-ASF is re-written as:
\begin{equation}
\begin{array}{l}
\begin{split}
    \overline{g}^{c}(\mathbf{x}|\mathbf{w})&=\max\limits_{1\leq i\leq m}\{(\overline{f_i}(\mathbf{x})-1)/{w_i}\}\\
    &+\alpha\sum\limits_{i=1}^{m}{(\overline{f_i}(\mathbf{x})-1)/{w_i}}
\end{split}
\end{array}.
\label{eq:nmaasf}
\end{equation}

\begin{figure}
	\centering
	\subfloat[Connvex PF]{\includegraphics[width=.4\linewidth]{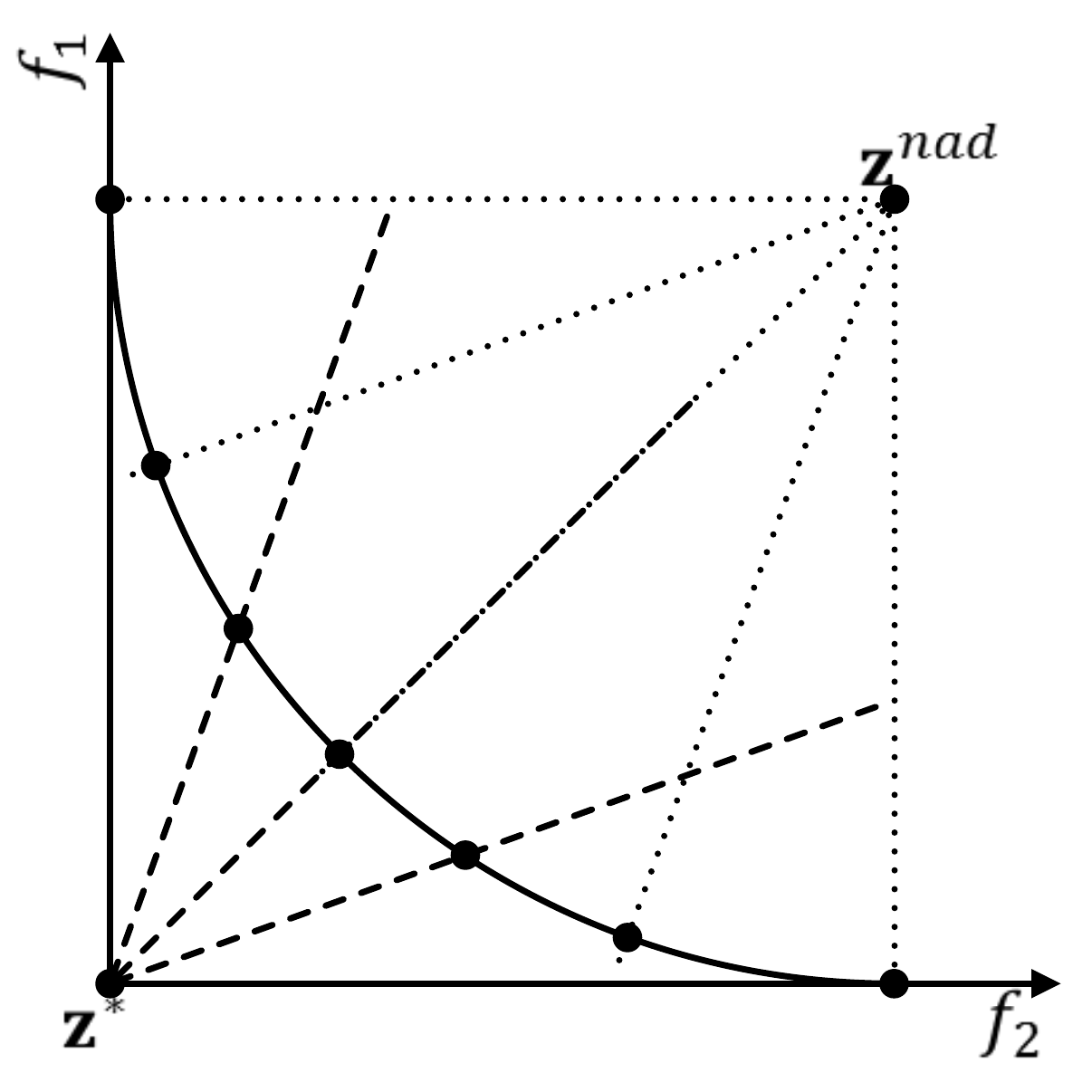}}
	\subfloat[Concave PF]{\includegraphics[width=.4\linewidth]{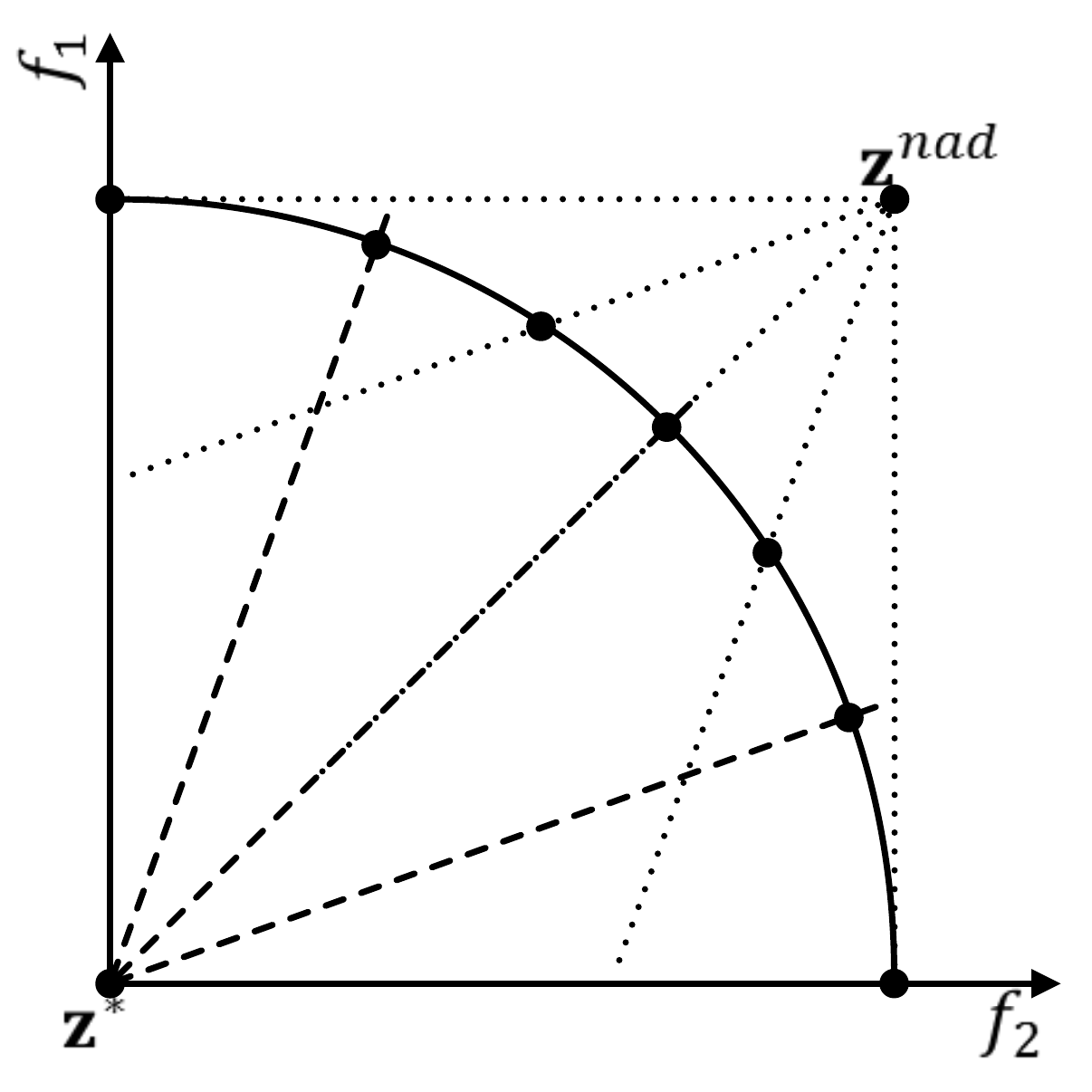}}
    \caption{Illustration of the complementary effects achieved by adversarial search directions on the convex and concave PFs.}
	\label{fig:weights}
\end{figure}

In the following paragraphs, we will discuss the complementary effects achieved by the PBI function and MA-ASF.
\begin{itemize}
    \item As shown in~\pref{fig:decomposition}(b) and~\pref{fig:decomposition}(c), the control areas of the PBI function with $\theta=5.0$ is much narrower than that of the MA-ASF. In this case, the population is driven towards the corresponding weight vector by using the PBI function as the selection criteria. Moreover, comparing to the MA-ASF, the control areas shared by different weight vectors are smaller in the PBI function. Accordingly, various weight vectors have a larger chance to hold different elite solutions and we can expect an improvement on the population diversity. On the other hand, due to the narrower control area, the selection pressure, with regard to the convergence, of the PBI function is not as strong as the MA-ASF. In other words, some solutions, which can update the subproblem maintained by a MA-ASF, might not be able to update the subproblem maintained by a PBI function. For example, as shown in~\pref{fig:decomposition}(b), although $\mathbf{F}^2\preceq\mathbf{F}^1$, the PBI function value of $\mathbf{F}^2$ is worse than that of $\mathbf{F}^1$ with regard to $\mathbf{w}^1$. In this case, the PBI function has a high risk of compromising the population convergence. 
    \item The other reason, which results in the different behaviors of the PBI function and MA-ASF, is their adversarial search directions by using the ideal point and nadir point respectively. More specifically, the PBI function pushes solutions toward the ideal point as close as possible; while the MA-ASF pushes the solutions away from the nadir point as far as possible. Therefore, given the same set of weight vectors, the search regions of the PBI function and MA-ASF complement each other. For example, for a convex PF shown in~\pref{fig:weights}(a), solutions found by the PBI function using the ideal point concentrate on the central region of the PF; in contrast, those found by the MA-ASF using the nadir point fill the gap between the central region and the boundary. As for a concave PF shown in~\pref{fig:weights}(b), solutions found by the PBI function using the ideal point sparsely spread over the PF; while those found by the MA-ASF using the nadir point improve the population diversity.
\end{itemize}

In summary, by using the scalarizing functions introduced above, i.e., the PBI function and the MA-ASF, the adversarial decomposition method makes the two co-evolving populations become complementary, i.e., one is diversity-oriented (denoted as the diversity population $S_d$) and the other is convergence-oriented (denoted as the convergence population $S_c$). In addition, the search regions is also diversified so that the solutions are able to cover a wider range of the PF.

\subsection{Population Update}
\label{sec:update}
\begin{algorithm}[!t]
	\caption{$\mathsf{PopulationUpdate}(S_c,S_d,\overline{\mathbf{x}},W)$}
	\label{alg:update}
	\KwIn{$S_c$, $S_d$, $W$ and an offspring solution $\overline{\mathbf{x}}$}
	\KwOut{Updated $S_c$ and $S_d$}
	\For{$i\leftarrow 1$ \KwTo $N$}{
		$D_d[i]\leftarrow d_2(\overline{F}(\overline{\mathbf{x}})|\mathbf{w}^i,\textbf{0})$;\\
	}
	$i_d\leftarrow \argmin\limits_{0\leq i\leq N}D_d[i]$\\
	\If{$\overline{g}^{d}(\overline{\mathbf{x}}|\mathbf{w}^{i_d}) <= \overline{g}^{d}(\mathbf{x}_d^{i_d}|\mathbf{w}^{i_d})$}{
		$\mathbf{x}_d^{i_d}\leftarrow \overline{\mathbf{x}}$;\\
	}
	\For{$i\leftarrow 1$ \KwTo $N$}{
		$D_c[i]\leftarrow d_2(\overline{F}(\overline{\mathbf{x}})|\mathbf{w}^i,\textbf{1})$;\\
	}
	$I_c\leftarrow$ Sort $D_c$ in ascending order and return the indexes;\\
	$t\leftarrow 0$;\\
	\For{$i\leftarrow 1$ \KwTo $N$}{
		\If{$\overline{g}^{c}(\overline{\mathbf{x}}|\mathbf{w}^{I_c[i]}) <= \overline{g}^{c}(\mathbf{x}_c^{I_c[i]}|\mathbf{w}^{I_c[i]})$}{
			$\mathbf{x}_c^{I_c[i]}\leftarrow \overline{\mathbf{x}}$, $t$++;\\
			$\mathbf{x}_c^{I_c[i]}.closeness\leftarrow i$, $\mathbf{x}_c^{I_c[i]}.closestP\leftarrow I_c[0]$;\\
			
			\If{$t==nr_c$} {
				\textbf{break};
			}
		}
	}
	
	\Return $S_c$, $S_d$;
\end{algorithm}

After the generation of an offspring solution $\overline{\mathbf{x}}$, it is used to update $S_d$ and $S_c$ separately. Note that the optimal solution for each subproblem is assumed to be along its corresponding reference line that connects the origin and the weight vector~\cite{LiKZD15}. Thus, to make $S_d$ as diversified as possible, we expect that different subproblems can hold different elite solutions. In this case, we restrict $\overline{\mathbf{x}}$ to only updating its closest subproblem (as shown in line 1 to line 5 of~\pref{alg:update}). As for $S_c$, its major purpose is to propel the population to the PF. To accelerate the convergence progress, we allow a dominantly superior solution to take over more than one subproblem, say $nr_c\geq1$. In particular, we at first sort the priorities of different subproblems according to their distances to $\overline{\mathbf{x}}$. It can update its $nr_c$ closest subproblems in case $\overline{\mathbf{x}}$ has a better MA-ASF function value (as shown line 6 to line 15 of~\pref{alg:update}). It is worth noting that we reserve two additional terms, in line 13 of~\pref{alg:update}, to facilitate the mating selection procedure introduced in~\pref{sec:mating}. One is the degree of closeness of the updated solution to its corresponding subproblem, denoted as $closeness$; the other is the index of this solution's closest subproblem, denoted as $closestP$.

\subsection{Mating Selection and Reproduction}
\label{sec:reproduction}

The interaction between the two co-evolving populations is an essential step in algorithms that consider multiple populations~\cite{LiKD15,ChenLiY17}. To take advantage of the complementary effects between $S_d$ and $S_c$, the interaction is implemented as a restricted mating selection mechanism that chooses the mating parents according to their working regions. Generally speaking, it contains two consecutive steps: one is the pairing step that makes each solution in $S_d$ be paired with a solution in $S_c$; the other is the mating selection step that selects the appropriate parents for the offspring reproduction. We will illustrate them in detail as follows.

\subsubsection{Pairing Step}
\label{sec:pairing}

To facilitate the latter mating selection step, the pairing step divides the two populations into different solution pairs, each of which contains two solutions from $S_d$ and $S_c$ respectively. This is achieved by finding a one-one stable matching between the solutions in $S_d$ and $S_c$. As a result, solutions in the same pair are regarded to have a similar working regions of the PF. 

To find a stable matching between solutions in $S_d$ and $S_c$, we need to define their mutual preferences at first. Specifically, since each solution in $S_d$ is close to its corresponding subproblem, we define the preference of a solution in $S_d$ (denoted as $\mathbf{x}_d$) to a solution in $S_c$ (denoted as $\mathbf{x}_c$) as:
\begin{equation}
    \Delta_{DC}(\mathbf{x}_d,\mathbf{x}_c)=\overline{g}^{d}(\mathbf{x}_c|\mathbf{w}_d),
\label{eq:preference_p}
\end{equation}
where $\mathbf{w}_d$ is the weight vector of the subproblem occupied by $\mathbf{x}_d$. As for the preference of $\mathbf{x}_c$ to $\mathbf{x}_d$, it is defined as:
\begin{equation}
\Delta_{CD}(\mathbf{x}_c,\mathbf{x}_d)=d_2(\overline{F}(\mathbf{x}_c)|\mathbf{w}_d,\textbf{0}).
\label{eq:preference_x}
\end{equation}

\begin{figure}[!t]
	\centering
	\subfloat{\includegraphics[width=.5\linewidth]{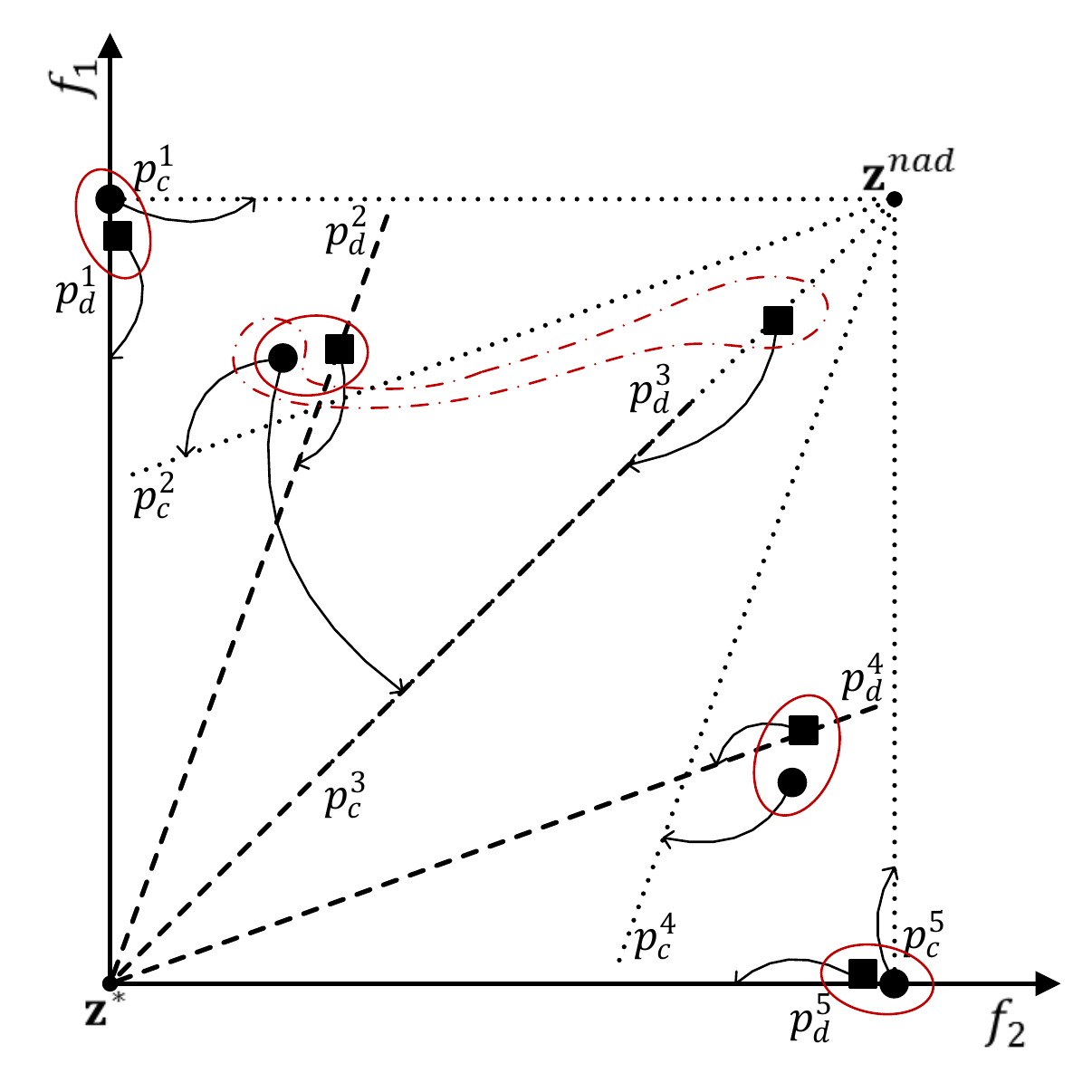}}
	\caption{Illustration of a pairing result. $p_d^i$ and $p_c^i$, $i\in\{1,\cdots,5\}$, indicate the subproblem in $S_d$ and $S_c$, respectively. The association of a solution and a subproblem is represented as a arrow.}
	\label{fig:match}
\end{figure}

Then, we sort the preference list of each solution in an ascending order and apply our recently developed two-level one-one stable matching method~\cite{STM2L,WuLKZZ17} to find a stable matching. Note that the two-level stable matching method is able to match each agent with one of its most preferred partners. Since a solution of an $m$-objective problem always locates within the local area between $m$ closest weight vectors, the length of the preference list is reduced to $m$ in the first-level matching process~\cite{WuLKZZ17}. As a result, the matched solutions in the first-level stable matching should work on the similar regions of the PF. For example, as shown in~\pref{fig:match}, the solution pairs formed in the first-level stable matching are surrounded by the red solid curves. From this figure, we can see that these matched solutions are close to each other and work on the similar regions. Therefore, we set the corresponding index of a sentinel array $R$ to 1, i.e., $R[i]=1$ where $i\in\{1,2,4,5\}$ and denote the corresponding subproblems have collaborative information. During the second-level matching process, the remaining solutions are matched based on the complete preference lists. Note that the matched solutions in the second-level stable matching are not guaranteed to work on the similar regions any longer. As shown in~\pref{fig:match}, the solution pair formed in second-level stable matching, surrounded by the red dotted curve, are away from each other. Thus, we set $R[3]=0$. The pseudo code of this pairing step is presented in~\pref{alg:stm2l}.

\begin{algorithm}[bt]
	\caption{$\mathsf{Match}(S_c,W)$}
	\label{alg:stm2l}
	\KwIn{$S_c$, $W$}
	\KwOut{Matching array $M$ and sentinel array $R$}
	Calculate preference lists for $S_c$ and $P_d$;\\
	Compute a two-level stable matching between $S_c$ and $P_d$;\\
	\For{$i\leftarrow 1$ \KwTo $N$}{
		$M[i]\leftarrow$ Index of the matching mate in $S_c$ of $\mathbf{x}_d^i$;\\
		\uIf{$\mathbf{x}_d^i$ finds a stable matching mate in the first-level stable matching}{
			$R[i]\leftarrow 1$\tcp*{the $i$-th subproblem has collaborative information}
		}
		\Else {
			$R[i]\leftarrow 0$;\\
		}
	}
	\Return $M$, $R$
\end{algorithm}

\subsubsection{Mating Selection Step}
\label{sec:mating}

The mating parents consist of two parts: one is the principal parent; the others are from its neighbors. Note that each solution pair is only allowed to contribute at most one mating parent to avoid wasting the computational resources upon the similar regions. Given a solution pair $(\mathbf{x}^i_d,\mathbf{x}^{M[i]}_c)$, the first step is to decide the population from which the principal parent is selected. This depends on the following three criteria. 
\begin{itemize}
    \item The first criterion is the \textit{subproblem's relative improvement}. As for $\mathbf{x}^i_d$, it is defined as:
	\begin{equation}
	\Delta_d^{i}=\frac{\overline{g}^{d}(\mathbf{x}_d^{i,old}|\mathbf{w}^i)-\overline{g}^{d}(\mathbf{x}_d^{i,new}|\mathbf{w}^i)}{\overline{g}^{d}(\mathbf{x}_d^{i,old}|\mathbf{w}^i)},
	\end{equation}
	where $\mathbf{x}_d^{i,new}$ and $\mathbf{x}_d^{i,old}$ are respectively the current and previous solution of the $i$th subproblem in $S_d$, respectively. As for $\mathbf{x}^{M[i]}_c$, it is defined as:
	\begin{equation}
	\Delta_c^{M[i]}=\left|\frac{\overline{g}^{c}(\mathbf{x}_c^{M[i],old}|\mathbf{w}^{M[i]})-\overline{g}^{c}(\mathbf{x}_c^{M[i],new}|\mathbf{w}^{M[i]})}{\overline{g}^{c}(\mathbf{x}_c^{M[i],old}|\mathbf{w}^{M[i]})}\right|.
	\end{equation}
	If $\Delta_d^i$ and $\Delta_c^{M[i]}$ are in a tie, we need the following two secondary criteria for decision making.
	\item As discussed in~\pref{sec:ad}, a solution found by the PBI function is not guaranteed to be non-dominated. If $\mathbf{x}_d^i$ is dominated by a solution in $S_d\cup S_c$, it is inappropriate to be chosen as a mating parent.
	\item As discussed in~\pref{sec:update}, a dominantly superior solution can occupy more than one subproblem by considering the MA-ASF. In this case, such solution may occupy a subproblem away from its working region, which makes it inadequate to be a mating parent.
\end{itemize}

\begin{algorithm}[bt]
	\caption{$\mathsf{PopSelection}(S_c,S_d,i,M,W)$}
	\label{alg:popSelect}
	\KwIn{$S_c$, $S_d$, $W$, matching array $M$ and the subproblem index $i$}
	\KwOut{population index $pop$}
	\uIf{$\Delta_d^i > \Delta_c^{M[i]}$}{
		$pop\leftarrow 1$\tcp*{chosen from $S_d$}
	}
	\uElseIf{$\Delta_d^i < \Delta_c^{M[i]}$}{
		$pop\leftarrow 2$\tcp*{chosen from $S_c$}
	}
	\Else{
		\uIf{$\mathbf{x}_d^i$ is nondominated \rm{\textbf{and}} $\mathbf{x}_c^{M[i]}.closeness > m$}{
			$pop\leftarrow 1$;\\
		}
		\uElseIf{$\mathbf{x}_d^i$ is dominated \rm{\textbf{and}} $\mathbf{x}_c^{M[i]}.closeness\leq m$}{
			$pop\leftarrow 2$;\\
		}
		\Else{
			$pop\leftarrow$ Randomly select from $\{1,2\}$;\\
		}
	}
	\Return $pop$;
\end{algorithm}

The pseudo code of the principal parent selection mechanism is given in~\pref{alg:popSelect}. If the subproblem's relative improvement of $\mathbf{x}_d^i$ is larger than that of $\mathbf{x}_c^{M[i]}$, it means that the corresponding subproblem of $\mathbf{x}_d^i$ has a higher potential for further improvement. And the principal parent should be selected from $S_d$, i.e., $\mathbf{x}_d^i$. Otherwise, $\mathbf{x}_c^{M[i]}$ will be chosen as the principal parent (line 1 to line 4 of~\pref{alg:popSelect}). If $\Delta_d^i$ and $\Delta_c^{M[i]}$ are in a tie, we need to check the domination status of $\mathbf{x}_d^i$ and $\mathbf{x}_c^{M[i]}$'s closeness to the corresponding subproblem (line 6 to line 11 of~\pref{alg:popSelect}). By comparing the subproblems' relative improvements, we can expect an efficient allocation of the computational resources to different regions of the PF. The two secondary criteria implicitly push the solutions towards the corresponding weight vectors thus improve the population diversity.

\begin{algorithm}[!t]
	\caption{$\mathsf{MatingSelection}(S_c,S_d,i,M,R,W,B)$}
	\label{alg:matingSelect}
	\KwIn{$S_c$, $S_d$, $W$, matching array $M$ and the subproblem index $i$, sentinel array $R$, neighborhood structure $B$}
	\KwOut{mating parents $\overline{S}$}
	$pop\leftarrow\mathsf{PopSelection}(S_c,S_d,i,M,W)$;\\
	\uIf{$rand<\delta$}{
		$S_p\leftarrow\emptyset$;\\
		\uIf{$pop==1$}{
			\For{$j\leftarrow 1$ \KwTo $T$}{
				$S_p\leftarrow S_p\cup\{\mathbf{x}_d^{B[i][j]}$\};\\
				\If{$R[B[i][j]]==1$}{
					$S_p\leftarrow S_p\cup\{\mathbf{x}_c^{M[B[i][j]]}$\};\\
				}
			}
			$\mathbf{x}^r\leftarrow$ Randomly select a solution from $S_p$;\\
			$\overline{S}\leftarrow\{\mathbf{x}_d^i,\mathbf{x}^r$\};\\
		}
		\Else{
			\For{$j\leftarrow 1$ \KwTo $T$}{
                \If{$\mathbf{x}_c^{B[M[i][j]]}.closestP\neq \mathbf{x}_c^{M[i]}.closestP$}{
					$S_p\leftarrow S_p\cup\{\mathbf{x}_c^{B[M[i]][j]}$\};\\
				}
			}
			$\mathbf{x}^r\leftarrow$ Randomly select a solution from $S_p$;\\
			$\overline{S}\leftarrow\{\mathbf{x}_c^{M[i]},\mathbf{x}^r$\};\\
		}
	}
	\Else{
		$S_p\leftarrow S_c\cup S_d$;\\
		$\mathbf{x}^r\leftarrow$ Randomly select a solution from $S_p$;\\
		\uIf{$pop==1$}{
			$\overline{S}\leftarrow\{\mathbf{x}_d^i,\mathbf{x}^r$\};\\
		}
		\Else{
			$\overline{S}\leftarrow\{\mathbf{x}_c^{M[i]},\mathbf{x}^r$\};\\	
		}
	}
	\Return $\overline{S}$
\end{algorithm}

After the selection of the principal parent, the other mating parent are selected from the neighbors of the subproblem occupied by the principal parent. More specifically, if the principal parent is from $S_d$, we store the solutions of its neighboring subproblems from both $S_d$ and $S_c$ into a temporary mating pool $S_p$. Note that only those subproblems having collaborative information (i.e., $R[i]=1$) are considered when choosing solutions from $S_c$ (line 5 and line 8 of~\pref{alg:matingSelect}). On the other hand, if the principal parent is from $S_c$, only solutions from $S_c$ have the chance to be stored in $S_p$. Note that we do not consider the solution that has the same closest subproblem as the principal parent (line 12 and line 14 of~\pref{alg:matingSelect}). Once $S_p$ is set up, the other mating parents are randomly chosen from it.

This paper uses the simulated binary crossover (SBX)~\cite{DebA94} and polynomial mutation~\cite{PolyMutation} for offspring generation. The mating parents are treated, while only one offspring solution will be randomly chosen for updating $S_d$ and $S_c$.


\section{Experimental Setup}
\label{sec:setup}

In this section, we describe the setup of our empirical studies, including the benchmark problems, performance indicator, peer algorithms and their parameter settings. 

\subsection{Benchmark Problems}
\label{sec:benchmark}

Here we choose DTLZ1 to DTLZ4~\cite{DTLZ}, WFG1 to WFG9~\cite{WFG}, and their minus version proposed in~\cite{IshibuchiSMN17}, i.e., DTLZ1$^{-1}$ to DTLZ4$^{-1}$ and WFG1$^{-1}$ to WFG9$^{-1}$ to form the benchmark problems in our empirical studies. In particular, the number of objectives are set as $m\in\{3,5,8,10,15\}$. The number of decision variables of DTLZ and DTLZ$^{-1}$ problem instances~\cite{DTLZ} is set to $n=m+k+1$, where $k=5$ for DTLZ1 and DTLZ1$^{-1}$ and $k=10$ for the others. As for WFG and WFG$^{-1}$ problem instances~\cite{WFG}, we set $n=k+l$, where $k=2\times(m-1)$ and $l=20$. Note that the DTLZ and WFG benchmark problems have been widely used for benchmarking the performance of many-objective optimizers; while their minus version is proposed to investigate the resilience to the irregular PF shapes. All these benchmark problems are scalable to any number of objectives.

\subsection{Performance Indicator}
\label{sec:indicator}

In our empirical studies, we choose the widely used Hypervolume (HV) indicator~\cite{HV} to quantitatively evaluate the performance of a many-objective optimizer. Specifically, the HV indicator is calculated as:
\begin{equation}
    HV(S)=\textsf{VOL}(\bigcup \limits_{\mathbf{x}\in S} [f_1(\mathbf{x}),z^r_1]\times\cdots\times [f_m(\mathbf{x}),z^r_m]),
\end{equation}
where $S$ is the solution set, $\mathbf{z}^r=(z_1^r,\cdots,z_m^r)$ is a point dominated by all Pareto-optimal objective vectors and \textsf{VOL} indicates the Lebesgue measure. In our empirical studies, we set $\mathbf{z}^r=(2,\cdots,2)^T$ and the objective vectors are normalized to $[0,1]$ before calculating the HV. The larger the HV value is, the better the quality of $S$ is for approximating the true PF. Each algorithm is run 31 times independently and the Wilcoxon's rank sum test at 5\% significant level is performed to show whether the peer algorithm is significantly outperformed by our proposed MOEA/AD. Note that we choose the population that has the higher HV value as the output of our proposed MOEA/AD. 

\subsection{Peer Algorithms}
\label{sec:peers}

Here we choose nine state-of-the-art many-objective optimizers to validate the competitiveness of our proposed MOEA/AD. These peer algorithms belong to different types, including two decomposition-based algorithms (MOEA/D~\cite{MOEAD} and Global WASF-GA~\cite{SaboridoRL16}), two Pareto-based algorithms (PICEA-g~\cite{WangPF13} and VaEA~\cite{XiangZLC16}), two indicator-based algorithms (HypE~\cite{BaderZ11} and KnEA~\cite{ZhangTJ15}), two algorithms that integrates the decomposition- and Pareto-based selection together (NSGA-III~\cite{DebJ14} and $\theta$-DEA~\cite{YuanXWY16}), and an improved two-archive-based algorithm (Two$\_$Arch2~\cite{WangJY15}). Some further comments upon the peer algorithms are listed as follows.
\begin{itemize}
    \item \textit{MOEA/D} uses the original PBI function with $\theta=5.0$ for the DTLZ and WFG problem instances. As for the DTLZ$^{-1}$ and WFG$^{-1}$ problem instances, it uses the inverted PBI function~\cite{Sato14} with $\theta=0.1$ as suggested in~\cite{IshibuchiSMN17}. Note that the inverted PBI function replaces the ideal point with the nadir point in equation~\pref{eq:pbi}.
	\item \textit{Global WASF-GA} is a decomposition-based algorithm that selects solutions to survive according to the rankings of solutions to each subproblem. Similar to MOEA/AD, it uses the ideal point and nadir point simultaneously in its scalarizing function. However, instead of maintaining two co-evolving populations, Global WASF-GA only has a single population, where half of it are maintained by the scalarizing function using the ideal point while the other are maintained by the nadir point.
    \item \textit{PICEA-g} co-evolves a set of target vectors sampled in the objective space, which can be regarded as a second population and is used to guide the environmental selection.
	\item \textit{Two$\_$Arch2} maintains two archives via indicator-based selection and Pareto-based selection separately. In particular, an $L_p$-norm-based diversity maintenance scheme is designed to maintain the diversity archive.
\end{itemize}

\subsection{Parameter Settings}
\begin{itemize}
    \item \textit{Weight vector}: We employ the method developed in~\cite{LiDZK15} to generate the weight vectors used in the MOEA/D variants. Note that we add an additional weight vector $\{1/m,\cdots,1/m\}$ to remedy the missing the centroid on the simplex for the 8-, 10- and 15-objective cases.
	\item \textit{Population size}: We set the population size the same as the number of weight vectors. In particular, $N$ is set as 91, 210, 157, 276 and 136 for $m=\{3,5,8,10,15\}$ respectively. 
    \item \textit{Termination criteria}: As suggested in~\cite{DebJ14}, the termination criterion is set as a predefined number of generations, as shown in~\pref{tab:evaluations}.
	\begin{table}[!t]
		\centering
		\caption{Settings of the Number of Generations.}
		\label{tab:evaluations}
		\begin{tabular}{c|ccccc}
			\hline
			Problem Instance & $m$=3 & $m$=5 & $m$=8 & $m$=10 & $m$=15 \\
			\hline
			DTLZ1, DTLZ1$^{-1}$ & 400   & 600   & 750   & 1,000 & 1,500 \\
			\hline
			DTLZ2, DTLZ2$^{-1}$ & 250   & 350   & 500   & 750   & 1,000 \\
			\hline
			DTLZ3, DTLZ3$^{-1}$ & 1,000 & 1,000 & 1,000 & 1,500 & 2,000 \\
			\hline
			DTLZ4, DTLZ4$^{-1}$ & 600   & 1,000 & 1,250 & 2,000 & 3,000 \\
			\hline
			WFG, WFG$^{-1}$ & 400   & 750   & 1,500 & 2,000 & 3,000 \\
			\hline
		\end{tabular}
	\end{table}
    \item \textit{Reproduction operators}: For the SBX, we set the crossover probability $p_c$ to 1.0 and the distribution index $\eta_c$ to 30~\cite{DebJ14}. As for the polynomial mutation, the probability $p_m$ and distribution index $\eta_m$ are set to be $1/n$ and $20$~\cite{DebJ14}, respectively.
	\item \textit{Neighborhood size}: $T=20$~\cite{MOEAD}.
	\item \textit{Probability to select in the neighborhood}: $\delta=0.9$~\cite{MOEAD}.
\end{itemize}

The intrinsic parameters of the other peer algorithms are set according to the recommendations in their original papers.


\section{Empirical Studies}
\label{sec:experiments}

In this section, we present and discuss the comparison results of our proposed MOEA/AD with the other state-of-the-art peer algorithms. The mean HV values are given in~\pref{tab:DTLZ_2} to~\pref{tab:DTLZ_v}, where the best results are highlighted in boldface with a gray background.

\subsection{Comparisons on DTLZ and WFG Problem Instances}
\label{sec:results-DTLZ-WFG} 

\begin{table*}[!t]
	\tiny
	\centering
	\caption{Comparison Results of MOEA/AD and 9 Peer Algorithms on DTLZ Problem Instances.}
	\label{tab:DTLZ_2}
	\begin{tabular}{c|c|cccccccccc}
		\hline
		Problem & $m$     & ~~PBI~~ & GWASF & PICEA-g & VaEA  & ~HypE~  & KnEA  & NSGA-III & $\theta$-DEA & TwoArch2 & MOEA/AD \\
		\hline
		& 3     & 7.785e+0$^\dagger$ & 7.134e+0$^\dagger$ & 7.562e+0$^\dagger$ & 7.764e+0$^\dagger$ & 7.774e+0$^\dagger$ & 7.307e+0$^\dagger$ & 7.786e+0 & 7.738e+0$^\dagger$ & 7.785e+0$^\dagger$ & \cellcolor[rgb]{0.851, 0.851, 0.851}\textbf{7.787e+0} \\
		& 5     & 3.197e+1$^\dagger$ & 1.965e+1$^\dagger$ & 3.191e+1$^\dagger$ & 3.194e+1$^\dagger$ & 3.186e+1$^\dagger$ & 2.939e+1$^\dagger$ & 3.197e+1$^\dagger$ & 3.197e+1$^\dagger$ & 3.196e+1$^\dagger$ & \cellcolor[rgb]{0.851, 0.851, 0.851}\textbf{3.197e+1} \\
		DTLZ1 & 8     & 2.560e+2$^\dagger$ & 7.026e+1$^\dagger$ & 2.482e+2$^\dagger$ & 2.559e+2$^\dagger$ & 1.973e+2$^\dagger$ & 1.739e+2$^\dagger$ & 2.560e+2 & 2.560e+2$^\dagger$ & 2.559e+2$^\dagger$ & \cellcolor[rgb]{0.851, 0.851, 0.851}\textbf{2.560e+2} \\
		& 10    & 1.024e+3$^\dagger$ & 4.761e+2$^\dagger$ & 1.010e+3$^\dagger$ & 1.024e+3$^\dagger$ & 9.573e+2$^\dagger$ & 8.514e+2$^\dagger$ & 1.024e+3$^\dagger$ & 1.024e+3$^\dagger$ & 1.024e+3$^\dagger$ & \cellcolor[rgb]{0.851, 0.851, 0.851}\textbf{1.024e+3} \\
		& 15    & 3.275e+4 & 1.092e+3$^\dagger$ & 2.815e+4$^\dagger$ & 3.275e+4$^\ddagger$ & 0.000e+0$^\dagger$ & 2.621e+4$^\dagger$ & \cellcolor[rgb]{0.851, 0.851, 0.851}\textbf{3.276e+4}$^\ddagger$ & 3.264e+4$^\ddagger$ & 3.275e+4 & 3.270e+4 \\
		\hline
		& 3     & \cellcolor[rgb]{0.851, 0.851, 0.851}\textbf{7.413e+0}$^\ddagger$ & 7.301e+0$^\dagger$ & 7.376e+0$^\dagger$ & 7.405e+0$^\dagger$ & 7.371e+0$^\dagger$ & 7.393e+0$^\dagger$ & 7.412e+0 & 7.413e+0$^\ddagger$ & 7.407e+0$^\dagger$ & 7.412e+0 \\
		& 5     & 3.170e+1 & 3.037e+1$^\dagger$ & 3.165e+1$^\dagger$ & 3.166e+1$^\dagger$ & 3.117e+1$^\dagger$ & 3.166e+1$^\dagger$ & 3.169e+1$^\dagger$ & 3.170e+1$^\dagger$ & 3.161e+1$^\dagger$ & \cellcolor[rgb]{0.851, 0.851, 0.851}\textbf{3.170e+1} \\
		DTLZ2 & 8     & 2.558e+2$^\dagger$ & 2.423e+2$^\dagger$ & 2.551e+2$^\dagger$ & 2.558e+2$^\dagger$ & 2.525e+2$^\dagger$ & 2.558e+2$^\dagger$ & 2.558e+2$^\dagger$ & 2.558e+2$^\dagger$ & 2.554e+2$^\dagger$ & \cellcolor[rgb]{0.851, 0.851, 0.851}\textbf{2.558e+2} \\
		& 10    & 1.024e+3$^\dagger$ & 6.398e+2$^\dagger$ & 1.023e+3$^\dagger$ & 1.024e+3$^\dagger$ & 1.017e+3$^\dagger$ & 1.024e+3$^\dagger$ & 1.024e+3$^\dagger$ & 1.024e+3$^\dagger$ & 1.022e+3$^\dagger$ & \cellcolor[rgb]{0.851, 0.851, 0.851}\textbf{1.024e+3} \\
		& 15    & 3.276e+4 & 1.638e+4$^\dagger$ & 3.231e+4$^\dagger$ & 3.276e+4 & 3.207e+4$^\dagger$ & \cellcolor[rgb]{0.851, 0.851, 0.851}\textbf{3.277e+4} & 3.276e+4 & 3.276e+4 & 3.275e+4$^\dagger$ & 3.276e+4 \\
		\hline
		& 3     & 7.406e+0$^\ddagger$ & 6.504e+0$^\dagger$ & 6.849e+0$^\dagger$ & 7.401e+0 & 7.285e+0$^\dagger$ & 7.236e+0$^\dagger$ & 7.406e+0 & 7.403e+0 & \cellcolor[rgb]{0.851, 0.851, 0.851}\textbf{7.413e+0}$^\ddagger$ & 7.403e+0 \\
		& 5     & 3.169e+1 & 1.600e+1$^\dagger$ & 2.997e+1$^\dagger$ & 3.166e+1$^\dagger$ & 1.332e+1$^\dagger$ & 2.862e+1$^\dagger$ & 3.169e+1$^\dagger$ & 3.169e+1$^\dagger$ & 3.159e+1$^\dagger$ & \cellcolor[rgb]{0.851, 0.851, 0.851}\textbf{3.169e+1} \\
		DTLZ3 & 8     & 2.459e+2$^\dagger$ & 1.278e+2$^\dagger$ & 2.081e+2$^\dagger$ & 2.237e+2$^\dagger$ & 0.000e+0$^\dagger$ & 1.542e+2$^\dagger$ & 2.558e+2 & 2.558e+2 & 2.542e+2$^\dagger$ & \cellcolor[rgb]{0.851, 0.851, 0.851}\textbf{2.558e+2} \\
		& 10    & 1.024e+3$^\dagger$ & 5.115e+2$^\dagger$ & 9.394e+2$^\dagger$ & 9.219e+2$^\dagger$ & 0.000e+0$^\dagger$ & 7.105e+2$^\dagger$ & 1.024e+3$^\dagger$ & 9.805e+2$^\dagger$ & 1.020e+3$^\dagger$ & \cellcolor[rgb]{0.851, 0.851, 0.851}\textbf{1.024e+3} \\
		& 15    & 2.328e+4$^\dagger$ & 1.631e+4$^\dagger$ & 2.422e+4$^\dagger$ & 2.674e+4$^\dagger$ & 0.000e+0$^\dagger$ & 0.000e+0$^\dagger$ & 3.276e+4 & 2.380e+4$^\dagger$ & 3.256e+4$^\dagger$ & \cellcolor[rgb]{0.851, 0.851, 0.851}\textbf{3.276e+4} \\
		\hline
		& 3     & 6.398e+0$^\dagger$ & 6.961e+0$^\dagger$ & 7.053e+0$^\dagger$ & 7.408e+0$^\dagger$ & \cellcolor[rgb]{0.851, 0.851, 0.851}\textbf{7.414e+0}$^\ddagger$ & 7.396e+0$^\dagger$ & 7.059e+0$^\dagger$ & 7.243e+0$^\dagger$ & 7.069e+0$^\dagger$ & 7.412e+0 \\
		& 5     & 3.087e+1$^\dagger$ & 2.710e+1$^\dagger$ & 3.144e+1$^\dagger$ & 3.167e+1$^\dagger$ & 3.149e+1$^\dagger$ & 3.167e+1$^\dagger$ & 3.170e+1$^\ddagger$ & \cellcolor[rgb]{0.851, 0.851, 0.851}\textbf{3.170e+1}$^\ddagger$ & 3.161e+1$^\dagger$ & 3.169e+1 \\
		DTLZ4 & 8     & 2.547e+2$^\dagger$ & 1.575e+2$^\dagger$ & 2.552e+2$^\dagger$ & 2.558e+2$^\dagger$ & 2.471e+2$^\dagger$ & 2.558e+2$^\dagger$ & 2.558e+2 & \cellcolor[rgb]{0.851, 0.851, 0.851}\textbf{2.558e+2} & 2.551e+2$^\dagger$ & 2.558e+2 \\
		& 10    & 1.023e+3$^\dagger$ & 7.165e+2$^\dagger$ & 1.023e+3$^\dagger$ & 1.024e+3$^\dagger$ & 1.019e+3$^\dagger$ & 1.024e+3$^\dagger$ & 1.024e+3 & 1.024e+3 & 1.022e+3$^\dagger$ & \cellcolor[rgb]{0.851, 0.851, 0.851}\textbf{1.024e+3} \\
		& 15    & 3.276e+4$^\dagger$ & 1.636e+4$^\dagger$ & 3.254e+4$^\dagger$ & 3.276e+4$^\dagger$ & 3.262e+4$^\dagger$ & \cellcolor[rgb]{0.851, 0.851, 0.851}\textbf{3.277e+4}$^\ddagger$ & 3.276e+4$^\dagger$ & 3.276e+4 & 3.274e+4$^\dagger$ & 3.276e+4 \\
		\hline
	\end{tabular}%
	\begin{tablenotes}
		\item[1] According to Wilcoxon's rank sum test, $^\dagger$ and $^\ddagger$ indicates whether the corresponding algorithm is significantly worse or better than MOEA/AD respectively.
	\end{tablenotes}
\end{table*}

Generally speaking, MOEA/AD is the most competitive algorithm for the DTLZ problem instances. As shown in~\pref{tab:DTLZ_2}, it wins in 161 out of 180 comparisons. More specifically, for DTLZ1 and DTLZ3, MOEA/AD obtains the largest HV values on all comparisons except for the 15-objective DTLZ1 and the 3-objective DTLZ3 instances. As for DTLZ2, MOEA/D-PBI obtains the best HV value on the 3-objective case, while MOEA/AD takes the leading position when the number of objectives becomes large. For DTLZ4, the best algorithm varies with different number of objectives. Even though MOEA/AD loses in 4 out of 45 comparisons, the differences are very slight. In addition, as for two decomposition-based algorithms, MOEA/D-PBI obtains significantly worse HV values than MOEA/AD on 14 out 20 comparisons, while Global WASF-GA were significantly outperformed on all 20 comparisons. In particular, Global WASF-GA fails to approximate the entire PF on all DTLZ instances due to its coarse diversity maintenance scheme. As for the two recently proposed Pareto-based many-objective optimizers, the HV values obtained by PICEA-g are significantly worse than MOEA/AD on all problem instances. This can be explained by its randomly sampled target vectors which slow down the convergence speed. VaEA performs slightly better than PICEA-g but it is still outperformed by MOEA/AD on 18 out of 20 comparisons. As expected, the performance of two indicator-based algorithms are not satisfied. In particular, KnEA merely obtains the best HV values on the 15-objective DTLZ2 and DTLZ4 instances. $\theta$-DEA and NSGA-III, which combine the decomposition- and Pareto-based selection methods together, achieve significantly better results than MOEA/AD in 2 and 3 comparisons respectively, where MOEA/AD beats them in 9 and 11 comparisons respectively. Two$\_$Arch2, which also maintains two co-evolving populations, is significantly outperformed by MOEA/AD on all DTLZ instances except for the 15-objective DTLZ1 and the 3-objective DTLZ3. Given these observations, we find that the genuine performance obtained by MOEA/AD does not merely come from the two co-evolving populations. The adversarial search directions and their collaborations help strike the balance between convergence and diversity.

\begin{table*}[!t]
	\renewcommand{\arraystretch}{0.8}
	\setlength\tabcolsep{5pt}
	\tiny
	\centering
	\caption{Comparison Results of MOEA/AD and 9 Peer Algorithms on WFG Problem Instances.}
	\label{tab:WFG_2}
	\begin{tabular}{c|c|cccccccccc}
		\hline
		Problem & m     & ~~PBI~~ & GWASF & PICEA-g & VaEA  & ~HypE~  & KnEA  & NSGA-III & $\theta$-DEA & TwoArch2 & MOEA/AD \\
		\hline
		& 3     & 5.515e+0$^\dagger$ & 6.031e+0$^\dagger$ & 5.277e+0$^\dagger$ & 5.317e+0$^\dagger$ & 3.739e+0$^\dagger$ & 5.380e+0$^\dagger$ & 4.455e+0$^\dagger$ & 5.317e+0$^\dagger$ & 6.287e+0$^\dagger$ & \cellcolor[rgb]{0.851, 0.851, 0.851}\textbf{6.355e+0} \\
		& 5     & 2.780e+1$^\dagger$ & 2.756e+1$^\dagger$ & 2.449e+1$^\dagger$ & 2.005e+1$^\dagger$ & 1.580e+1$^\dagger$ & 2.050e+1$^\dagger$ & 1.737e+1$^\dagger$ & 2.513e+1$^\dagger$ & 2.621e+1$^\dagger$ & \cellcolor[rgb]{0.851, 0.851, 0.851}\textbf{2.988e+1} \\
		WFG1  & 8     & 2.165e+2$^\dagger$ & 1.547e+2$^\dagger$ & 2.172e+2$^\dagger$ & 2.172e+2$^\dagger$ & 1.205e+2$^\dagger$ & 1.722e+2$^\dagger$ & 1.252e+2$^\dagger$ & 2.314e+2 & \cellcolor[rgb]{0.851, 0.851, 0.851}\textbf{2.454e+2}$^\ddagger$ & 2.330e+2 \\
		& 10    & 8.402e+2$^\dagger$ & 7.059e+2$^\dagger$ & 9.725e+2$^\ddagger$ & 9.434e+2$^\dagger$ & 4.828e+2$^\dagger$ & 7.520e+2$^\dagger$ & 5.228e+2$^\dagger$ & 9.562e+2$^\dagger$ & \cellcolor[rgb]{0.851, 0.851, 0.851}\textbf{1.002e+3}$^\ddagger$ & 9.673e+2 \\
		& 15    & 2.000e+4$^\dagger$ & 1.563e+4$^\dagger$ & 2.746e+4$^\ddagger$ & 3.040e+4$^\ddagger$ & 1.374e+4$^\dagger$ & 2.423e+4$^\dagger$ & 1.981e+4$^\dagger$ & 2.263e+4$^\dagger$ & \cellcolor[rgb]{0.851, 0.851, 0.851}\textbf{3.175e+4}$^\ddagger$ & 2.645e+4 \\
		\hline
		& 3     & 6.832e+0$^\dagger$ & 7.381e+0 & 7.545e+0$^\ddagger$ & 7.445e+0$^\ddagger$ & 7.077e+0 & 7.519e+0$^\ddagger$ & 7.423e+0$^\ddagger$ & 7.557e+0$^\ddagger$ & \cellcolor[rgb]{0.851, 0.851, 0.851}\textbf{7.637e+0}$^\ddagger$ & 7.186e+0 \\
		& 5     & 2.831e+1$^\dagger$ & 3.083e+1$^\ddagger$ & 3.179e+1$^\ddagger$ & 3.135e+1$^\ddagger$ & 3.041e+1 & 3.166e+1$^\ddagger$ & 3.127e+1$^\ddagger$ & \cellcolor[rgb]{0.851, 0.851, 0.851}\textbf{3.093e+1} & 3.181e+1$^\ddagger$ & 2.954e+1 \\
		WFG2  & 8     & 2.231e+2$^\dagger$ & 1.302e+2$^\dagger$ & 2.497e+2$^\ddagger$ & 2.503e+2$^\ddagger$ & 2.388e+2$^\dagger$ & 2.543e+2$^\ddagger$ & 2.471e+2$^\ddagger$ & 2.359e+2$^\dagger$ & \cellcolor[rgb]{0.851, 0.851, 0.851}\textbf{2.558e+2}$^\ddagger$ & 2.435e+2 \\
		& 10    & 9.246e+2$^\dagger$ & 5.203e+2$^\dagger$ & 1.010e+3$^\ddagger$ & 1.019e+3$^\ddagger$ & 9.801e+2$^\dagger$ & 1.019e+3$^\ddagger$ & 1.011e+3$^\ddagger$ & 9.597e+2$^\dagger$ & \cellcolor[rgb]{0.851, 0.851, 0.851}\textbf{1.024e+3}$^\ddagger$ & 9.853e+2 \\
		& 15    & 2.949e+4$^\dagger$ & 1.634e+4$^\dagger$ & 3.124e+4 & 3.195e+4$^\ddagger$ & 3.000e+4$^\dagger$ & 3.152e+4$^\dagger$ & 3.038e+4 & 2.592e+4$^\dagger$ & \cellcolor[rgb]{0.851, 0.851, 0.851}\textbf{3.276e+4}$^\ddagger$ & 3.133e+4 \\
		\hline
		& 3     & 6.348e+0$^\dagger$ & 7.010e+0$^\ddagger$ & 6.988e+0$^\ddagger$ & 6.818e+0$^\dagger$ & 6.608e+0$^\dagger$ & 6.870e+0 & 6.889e+0$^\dagger$ & 6.931e+0$^\ddagger$ & \cellcolor[rgb]{0.851, 0.851, 0.851}\textbf{7.048e+0}$^\ddagger$ & 6.906e+0 \\
		& 5     & 2.484e+1$^\dagger$ & 2.763e+1$^\dagger$ & 2.848e+1$^\ddagger$ & 2.666e+1$^\dagger$ & 2.695e+1$^\dagger$ & 2.615e+1$^\dagger$ & 2.753e+1$^\dagger$ & 2.810e+1$^\dagger$ & \cellcolor[rgb]{0.851, 0.851, 0.851}\textbf{2.860e+1}$^\ddagger$ & 2.835e+1 \\
		WFG3  & 8     & 1.559e+2$^\dagger$ & 1.278e+2$^\dagger$ & 2.222e+2 & 2.165e+2$^\dagger$ & 2.136e+2$^\dagger$ & 2.065e+2$^\dagger$ & 1.711e+2$^\dagger$ & 1.542e+2$^\dagger$ & \cellcolor[rgb]{0.851, 0.851, 0.851}\textbf{2.294e+2}$^\ddagger$ & 2.219e+2 \\
		& 10    & 4.934e+2$^\dagger$ & 5.114e+2$^\dagger$ & 9.010e+2$^\ddagger$ & 8.670e+2$^\dagger$ & 8.718e+2$^\dagger$ & 8.409e+2$^\dagger$ & 6.584e+2$^\dagger$ & 6.066e+2$^\dagger$ & \cellcolor[rgb]{0.851, 0.851, 0.851}\textbf{9.240e+2}$^\ddagger$ & 8.883e+2 \\
		& 15    & 1.340e+4$^\dagger$ & 1.634e+4$^\dagger$ & 2.772e+4 & 2.784e+4 & 2.700e+4$^\dagger$ & 2.141e+4$^\dagger$ & 1.964e+4$^\dagger$ & 1.835e+4$^\dagger$ & \cellcolor[rgb]{0.851, 0.851, 0.851}\textbf{2.946e+4}$^\ddagger$ & 2.774e+4 \\
		\hline
		& 3     & 7.191e+0$^\dagger$ & 7.246e+0$^\dagger$ & 7.307e+0$^\dagger$ & 7.282e+0$^\dagger$ & 7.088e+0$^\dagger$ & 7.324e+0$^\dagger$ & 7.317e+0$^\dagger$ & 7.319e+0$^\dagger$ & \cellcolor[rgb]{0.851, 0.851, 0.851}\textbf{7.372e+0}$^\ddagger$ & 7.369e+0 \\
		& 5     & 3.063e+1$^\dagger$ & 3.009e+1$^\dagger$ & 3.135e+1$^\dagger$ & 3.076e+1$^\dagger$ & 2.914e+1$^\dagger$ & 3.125e+1$^\dagger$ & 3.107e+1$^\dagger$ & 3.110e+1$^\dagger$ & 3.134e+1$^\dagger$ & \cellcolor[rgb]{0.851, 0.851, 0.851}\textbf{3.150e+1} \\
		WFG4  & 8     & 2.170e+2$^\dagger$ & 1.284e+2$^\dagger$ & 2.381e+2$^\dagger$ & 2.513e+2$^\dagger$ & 2.228e+2$^\dagger$ & 2.553e+2$^\dagger$ & 2.514e+2$^\dagger$ & 2.520e+2$^\dagger$ & 2.539e+2$^\dagger$ & \cellcolor[rgb]{0.851, 0.851, 0.851}\textbf{2.557e+2} \\
		& 10    & 8.608e+2$^\dagger$ & 5.129e+2$^\dagger$ & 9.697e+2$^\dagger$ & 1.006e+3$^\dagger$ & 9.242e+2$^\dagger$ & 1.022e+3$^\dagger$ & 1.010e+3$^\dagger$ & 1.013e+3$^\dagger$ & 1.018e+3$^\dagger$ & \cellcolor[rgb]{0.851, 0.851, 0.851}\textbf{1.024e+3} \\
		& 15    & 2.577e+4$^\dagger$ & 1.638e+4$^\dagger$ & 2.929e+4$^\dagger$ & 3.249e+4$^\dagger$ & 2.969e+4$^\dagger$ & 3.273e+4$^\dagger$ & 3.268e+4$^\dagger$ & 3.270e+4$^\dagger$ & 3.266e+4$^\dagger$ & \cellcolor[rgb]{0.851, 0.851, 0.851}\textbf{3.274e+4} \\
		\hline
		& 3     & 7.067e+0$^\dagger$ & 7.070e+0$^\dagger$ & 7.090e+0$^\dagger$ & 7.127e+0$^\dagger$ & 6.963e+0$^\dagger$ & \cellcolor[rgb]{0.851, 0.851, 0.851}\textbf{7.168e+0} & 7.130e+0$^\dagger$ & 7.132e+0$^\dagger$ & 7.148e+0 & 7.161e+0 \\
		& 5     & 3.026e+1$^\dagger$ & 2.940e+1$^\dagger$ & 3.044e+1$^\dagger$ & 3.024e+1$^\dagger$ & 2.940e+1$^\dagger$ & 3.064e+1$^\dagger$ & 3.049e+1$^\dagger$ & 3.053e+1$^\dagger$ & 3.049e+1$^\dagger$ & \cellcolor[rgb]{0.851, 0.851, 0.851}\textbf{3.073e+1} \\
		WFG5  & 8     & 2.310e+2$^\dagger$ & 1.816e+2$^\dagger$ & 2.400e+2$^\dagger$ & 2.459e+2$^\dagger$ & 2.146e+2$^\dagger$ & 2.470e+2$^\dagger$ & 2.460e+2$^\dagger$ & 2.459e+2$^\dagger$ & 2.451e+2$^\dagger$ & \cellcolor[rgb]{0.851, 0.851, 0.851}\textbf{2.470e+2} \\
		& 10    & 8.937e+2$^\dagger$ & 4.735e+2$^\dagger$ & 9.613e+2$^\dagger$ & 9.817e+2$^\dagger$ & 9.214e+2$^\dagger$ & 9.866e+2$^\dagger$ & 9.835e+2$^\dagger$ & 9.829e+2$^\dagger$ & 9.801e+2$^\dagger$ & \cellcolor[rgb]{0.851, 0.851, 0.851}\textbf{9.867e+2} \\
		& 15    & 2.515e+4$^\dagger$ & 1.509e+4$^\dagger$ & 2.924e+4$^\dagger$ & 3.138e+4$^\dagger$ & 2.922e+4$^\dagger$ & 3.142e+4 & 3.141e+4$^\dagger$ & 3.057e+4$^\dagger$ & 3.110e+4$^\dagger$ & \cellcolor[rgb]{0.851, 0.851, 0.851}\textbf{3.142e+4} \\
		\hline
		& 3     & 7.053e+0$^\dagger$ & 7.199e+0 & 7.203e+0 & 7.169e+0$^\dagger$ & 6.999e+0$^\dagger$ & 7.214e+0 & 7.179e+0$^\dagger$ & 7.190e+0$^\dagger$ & \cellcolor[rgb]{0.851, 0.851, 0.851}\textbf{7.238e+0} & 7.218e+0 \\
		& 5     & 2.896e+1$^\dagger$ & 2.950e+1$^\dagger$ & \cellcolor[rgb]{0.851, 0.851, 0.851}\textbf{3.078e+1} & 3.038e+1$^\dagger$ & 2.990e+1$^\dagger$ & 3.078e+1 & 3.061e+1 & 3.062e+1 & 3.074e+1 & 3.075e+1 \\
		WFG6  & 8     & 1.908e+2$^\dagger$ & 2.478e+2 & 2.466e+2$^\dagger$ & 2.476e+2 & 2.373e+2$^\dagger$ & \cellcolor[rgb]{0.851, 0.851, 0.851}\textbf{2.481e+2} & 2.470e+2$^\dagger$ & 2.475e+2 & 2.475e+2 & 2.479e+2 \\
		& 10    & 7.398e+2$^\dagger$ & 6.716e+2$^\dagger$ & 9.906e+2$^\dagger$ & 9.883e+2$^\dagger$ & 9.618e+2$^\dagger$ & 9.916e+2 & 9.880e+2$^\dagger$ & 9.901e+2$^\dagger$ & 9.885e+2$^\dagger$ & \cellcolor[rgb]{0.851, 0.851, 0.851}\textbf{9.941e+2} \\
		& 15    & 1.735e+4$^\dagger$ & 1.577e+4$^\dagger$ & 3.060e+4$^\dagger$ & 3.166e+4 & 3.012e+4$^\dagger$ & 3.163e+4 & 3.161e+4 & 3.115e+4$^\dagger$ & \cellcolor[rgb]{0.851, 0.851, 0.851}\textbf{3.172e+4} & 3.159e+4 \\
		\hline
		& 3     & 7.091e+0$^\dagger$ & 7.298e+0$^\dagger$ & 7.350e+0$^\dagger$ & 7.330e+0$^\dagger$ & 6.837e+0$^\dagger$ & 7.383e+0$^\dagger$ & 7.353e+0$^\dagger$ & 7.363e+0$^\dagger$ & \cellcolor[rgb]{0.851, 0.851, 0.851}\textbf{7.397e+0}$^\ddagger$ & 7.388e+0 \\
		& 5     & 3.030e+1$^\dagger$ & 3.036e+1$^\dagger$ & 3.150e+1$^\dagger$ & 3.113e+1$^\dagger$ & 2.872e+1$^\dagger$ & 3.160e+1$^\dagger$ & 3.140e+1$^\dagger$ & 3.148e+1$^\dagger$ & 3.157e+1$^\dagger$ & \cellcolor[rgb]{0.851, 0.851, 0.851}\textbf{3.163e+1} \\
		WFG7  & 8     & 2.028e+2$^\dagger$ & 1.912e+2$^\dagger$ & 2.491e+2$^\dagger$ & 2.545e+2$^\dagger$ & 2.236e+2$^\dagger$ & 2.556e+2$^\dagger$ & 2.542e+2$^\dagger$ & 2.546e+2$^\dagger$ & 2.553e+2$^\dagger$ & \cellcolor[rgb]{0.851, 0.851, 0.851}\textbf{2.558e+2} \\
		& 10    & 8.226e+2$^\dagger$ & 5.120e+2$^\dagger$ & 1.006e+3$^\dagger$ & 1.019e+3$^\dagger$ & 9.682e+2$^\dagger$ & 1.023e+3$^\dagger$ & 1.019e+3$^\dagger$ & 1.020e+3$^\dagger$ & 1.023e+3$^\dagger$ & \cellcolor[rgb]{0.851, 0.851, 0.851}\textbf{1.024e+3} \\
		& 15    & 1.813e+4$^\dagger$ & 1.638e+4$^\dagger$ & 3.095e+4$^\dagger$ & 3.272e+4$^\dagger$ & 3.024e+4$^\dagger$ & 3.233e+4$^\dagger$ & 3.267e+4$^\dagger$ & 3.153e+4$^\dagger$ & 3.275e+4$^\dagger$ & \cellcolor[rgb]{0.851, 0.851, 0.851}\textbf{3.276e+4} \\
		\hline
		& 3     & 6.989e+0$^\dagger$ & 7.096e+0$^\ddagger$ & 7.062e+0 & 7.069e+0 & 6.623e+0$^\dagger$ & 7.124e+0$^\ddagger$ & 7.082e+0 & 7.073e+0 & \cellcolor[rgb]{0.851, 0.851, 0.851}\textbf{7.185e+0}$^\ddagger$ & 7.082e+0 \\
		& 5     & 2.870e+1$^\dagger$ & 2.958e+1$^\dagger$ & 3.072e+1$^\dagger$ & 2.988e+1$^\dagger$ & 2.775e+1$^\dagger$ & 3.061e+1$^\dagger$ & 3.042e+1$^\dagger$ & 3.043e+1$^\dagger$ & 3.081e+1$^\dagger$ & \cellcolor[rgb]{0.851, 0.851, 0.851}\textbf{3.115e+1} \\
		WFG8  & 8     & 1.425e+2$^\dagger$ & 1.277e+2$^\dagger$ & 2.477e+2$^\dagger$ & 2.431e+2$^\dagger$ & 2.230e+2$^\dagger$ & 2.519e+2$^\dagger$ & 2.468e+2$^\dagger$ & 2.473e+2$^\dagger$ & 2.515e+2$^\dagger$ & \cellcolor[rgb]{0.851, 0.851, 0.851}\textbf{2.542e+2} \\
		& 10    & 5.725e+2$^\dagger$ & 4.530e+2$^\dagger$ & 9.986e+2$^\dagger$ & 9.773e+2$^\dagger$ & 9.496e+2$^\dagger$ & 1.015e+3$^\dagger$ & 9.980e+2$^\dagger$ & 9.989e+2$^\dagger$ & 1.016e+3$^\dagger$ & \cellcolor[rgb]{0.851, 0.851, 0.851}\textbf{1.021e+3} \\
		& 15    & 1.887e+4$^\dagger$ & 1.476e+4$^\dagger$ & 3.075e+4$^\dagger$ & 3.235e+4$^\dagger$ & 2.997e+4$^\dagger$ & 3.022e+4$^\dagger$ & 3.250e+4$^\dagger$ & 2.827e+4$^\dagger$ & 3.264e+4$^\dagger$ & \cellcolor[rgb]{0.851, 0.851, 0.851}\textbf{3.266e+4} \\
		\hline
		& 3     & 6.749e+0$^\dagger$ & 6.796e+0$^\dagger$ & 6.869e+0$^\dagger$ & 7.002e+0 & 6.725e+0$^\dagger$ & \cellcolor[rgb]{0.851, 0.851, 0.851}\textbf{7.032e+0} & 6.971e+0 & 6.985e+0 & 7.017e+0 & 7.014e+0 \\
		& 5     & 2.885e+1 & 2.799e+1$^\dagger$ & 2.900e+1$^\dagger$ & 2.884e+1$^\dagger$ & 2.730e+1$^\dagger$ & \cellcolor[rgb]{0.851, 0.851, 0.851}\textbf{2.981e+1} & 2.923e+1$^\dagger$ & 2.935e+1 & 2.960e+1 & 2.928e+1 \\
		WFG9  & 8     & 2.021e+2$^\dagger$ & 2.322e+2$^\ddagger$ & 2.304e+2$^\dagger$ & 2.318e+2$^\dagger$ & 2.046e+2$^\dagger$ & \cellcolor[rgb]{0.851, 0.851, 0.851}\textbf{2.434e+2}$^\ddagger$ & 2.317e+2$^\dagger$ & 2.346e+2 & 2.351e+2 & 2.330e+2 \\
		& 10    & 7.080e+2$^\dagger$ & 5.795e+2$^\dagger$ & 9.200e+2$^\dagger$ & 9.264e+2$^\dagger$ & 8.529e+2$^\dagger$ & \cellcolor[rgb]{0.851, 0.851, 0.851}\textbf{9.826e+2}$^\ddagger$ & 9.350e+2$^\dagger$ & 9.360e+2$^\dagger$ & 9.410e+2 & 9.376e+2 \\
		& 15    & 1.685e+4$^\dagger$ & 1.500e+4$^\dagger$ & 2.884e+4$^\dagger$ & 2.937e+4$^\dagger$ & 2.731e+4$^\dagger$ & \cellcolor[rgb]{0.851, 0.851, 0.851}\textbf{3.069e+4}$^\ddagger$ & 2.911e+4$^\dagger$ & 2.931e+4 & 2.884e+4$^\dagger$ & 2.957e+4 \\
		\hline
	\end{tabular}%
	\begin{tablenotes}
		\item[1] According to Wilcoxon's rank sum test, $^\dagger$ and $^\ddagger$ indicates whether the corresponding algorithm is significantly worse or better than MOEA/AD respectively.
	\end{tablenotes}
\end{table*}

\begin{figure*}[!t]
	\centering
	\subfloat[]{\includegraphics[width=.5\linewidth]{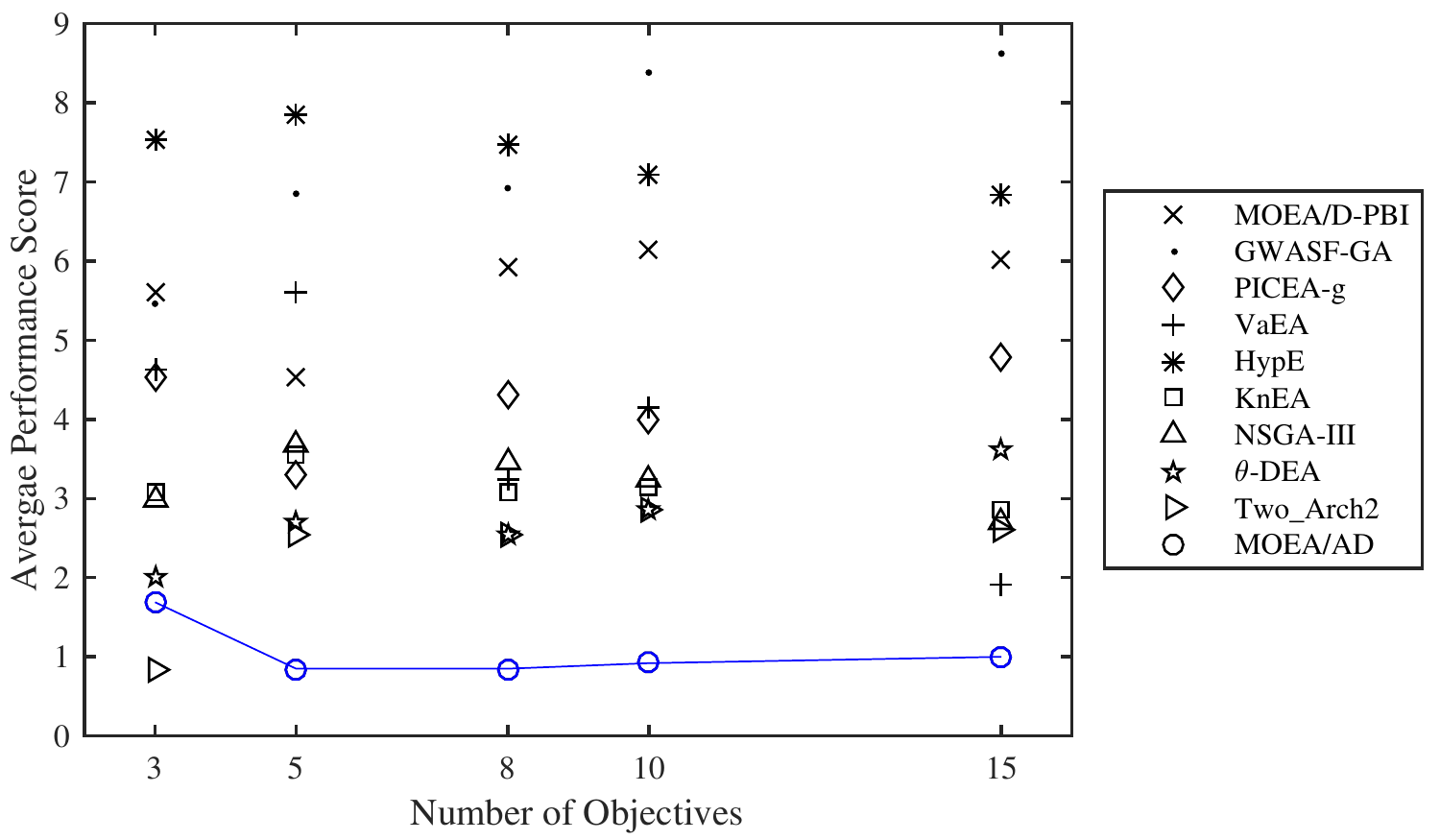}}
	\subfloat[]{\includegraphics[width=.5\linewidth]{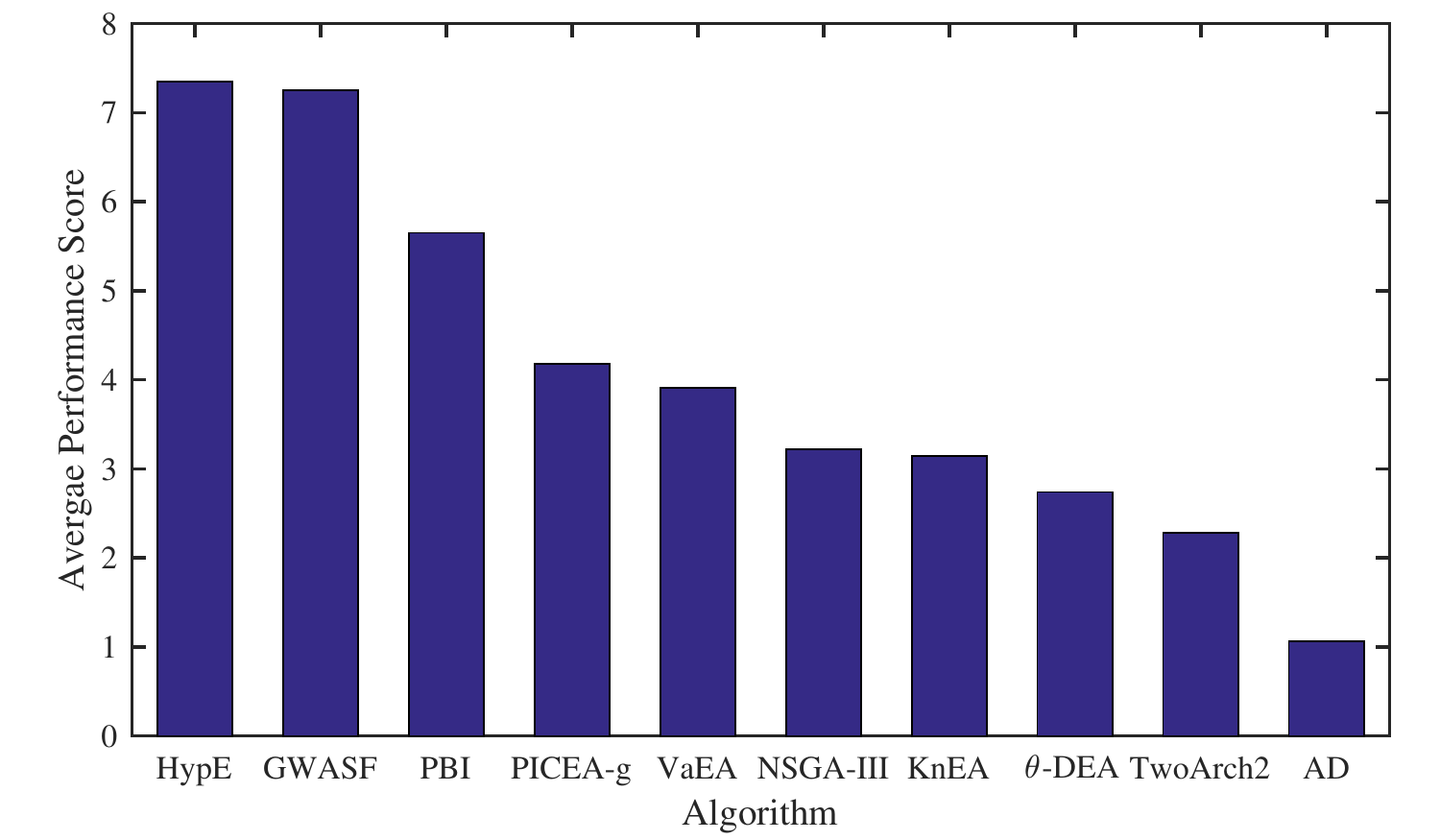}}
	\caption{(a) Average performance scores on different number of objectives over DTLZ and WFG problem instances. (b) Average performance scores over all DTLZ and WFG problem instances.}
	\label{fig:pscore}
\end{figure*}

The comparison results on WFG problem instances are given in \pref{tab:WFG_2}. From these results, we can see that MOEA/AD shows the best performance on all WFG4, WFG5, WFG7 and WFG8 instances when the number of objectives is larger than 3, where it wins in 143 out of 144 comparisons. For WFG6, although MOEA/AD only wins on the 10-objective case, its worse performance on the other WFG6 instances are not statistically significant. KnEA obtains the largest HV values on all WFG9 instances. As for WFG1 to WFG3, which have irregular PFs, Two$\_$Arch2 shows the best performance on 12 instances. In contrast, MOEA/AD only obtains the best HV results on 3- and 5-objective WFG1 instances. It is worth noting that NSGA-III, KnEA, PICEA-g and VaEA perform quite well on the WFG2 instances. Due to irregular PFs, the decomposition-based algorithms struggle to find a nondominated solution for each subproblem, while Two$\_$Arch2 together with the above four algorithms are still able to allocate all computational resources upon the PF. Nevertheless, HypE cannot perform as good as KnEA. Talking about the other algorithms, WASF-GA and MOEA/D-PBI give the worst overall performance. Unlike on the DTLZ problem instances, the performance MOEA/D-PBI degrades a lot on the WFG instances. In contrast, by simultaneously maintaining two complementary populations, the overall performance of our proposed MOEA/AD remains the best on the WFG instances.

To have a better overall comparison among different algorithms, we employ the performance score proposed in~\cite{BaderZ11} to rank the performance of different algorithms over different types of problem instances. Given $K$ algorithms, i.e. $A_1,\cdots,A_K$, the performance score of an algorithm $A_i,i\in\{1,\cdots,K\}$, is defined as
\begin{equation}
P(A_i)=\sum_{j=1,j\neq i}^{K}\delta_{i,j},
\end{equation}
where $\delta_{i,j}=1$ if $A_j$ is significant better than $A_i$ according to the Wilcoxon's rank sum test; otherwise, $\delta_{i,j}=0$. In other words, the performance score of an algorithm counts the number of competitors that outperform it on a given problem instance. Thus, the smaller performance score, the better an algorithm performs. The average performance scores of different algorithms on DTLZ and WFG problem instances are shown in \pref{fig:pscore}. From \pref{fig:pscore}(a), we find that MOEA/AD is the best algorithm on problems with more than 3 objectives and its better scores are of statistical significance. Two$\_$Arch2 shows the best performance on the 3-objective cases, but its performance significantly degenerates with the dimensionality. By aggregating the average performance scores on all problem instances, \pref{fig:pscore}(b) demonstrates the comparisons on all DTLZ and WFG problem instances. Similar to the previous observation, followed by $\theta$-DEA and Two$\_$Arch2, our proposed MOEA/AD obtains the best overall performance.

\subsection{Comparisons on DTLZ$^{-1}$ and WFG$^{-1}$ Problem Instances}
\label{sec:results-inverse}

\begin{table*}[!t]
	\renewcommand{\arraystretch}{0.8}
	\setlength\tabcolsep{5pt}
	\tiny
	\centering
	\caption{Comparison Results of MOEA/AD and 9 Peer Algorithms on DTLZ$^{-1}$ Problem Instances.}
	\label{tab:iDTLZ_2}
	\begin{tabular}{cccccccccccc}
		\hline
		Problem & m     & ~~IPBI~~ & GWASF & PICEA-g & VaEA  & ~HypE~  & KnEA  & NSGA-III & $\theta$-DEA & TwoArch2 & MOEA/AD \\
		\hline
		& 3     & 5.048e+0$^\dagger$ & 5.357e+0$^\dagger$ & 3.840e+0$^\dagger$ & 5.317e+0$^\dagger$ & 5.344e+0$^\dagger$ & 4.659e+0$^\dagger$ & 5.318e+0$^\dagger$ & 5.257e+0$^\dagger$ & 5.361e+0$^\dagger$ & \cellcolor[rgb]{0.851, 0.851, 0.851}\textbf{5.394e+0} \\
		& 5     & 8.696e+0$^\dagger$ & 1.021e+1$^\dagger$ & 5.550e+0$^\dagger$ & 1.016e+1$^\dagger$ & 1.016e+1$^\dagger$ & 7.865e+0$^\dagger$ & 8.780e+0$^\dagger$ & 6.791e+0$^\dagger$ & 9.926e+0$^\dagger$ & \cellcolor[rgb]{0.851, 0.851, 0.851}\textbf{1.088e+1} \\
		DTLZ1$^{-1}$ & 8     & 1.765e+1$^\dagger$ & 8.880e+0$^\dagger$ & 6.900e+0$^\dagger$ & 1.883e+1$^\dagger$ & 1.838e+1$^\dagger$ & 6.773e+0$^\dagger$ & 6.640e+0$^\dagger$ & 2.538e+0$^\dagger$ & 1.359e+1$^\dagger$ & \cellcolor[rgb]{0.851, 0.851, 0.851}\textbf{1.949e+1} \\
		& 10    & 2.600e+1$^\dagger$ & 1.119e+1$^\dagger$ & 8.024e+0$^\dagger$ & 2.815e+1$^\dagger$ & 2.305e+1$^\dagger$ & 1.183e+1$^\dagger$ & 7.692e+0$^\dagger$ & 2.497e+0$^\dagger$ & 1.586e+1$^\dagger$ & \cellcolor[rgb]{0.851, 0.851, 0.851}\textbf{2.946e+1} \\
		& 15    & 4.124e+1 & 6.375e+0$^\dagger$ & 6.842e+0$^\dagger$ & 3.918e+1$^\dagger$ & 1.652e+1$^\dagger$ & 1.079e+1$^\dagger$ & 6.047e+0$^\dagger$ & 2.508e+0$^\dagger$ & 1.905e+1$^\dagger$ & \cellcolor[rgb]{0.851, 0.851, 0.851}\textbf{4.125e+1} \\
		\hline
		& 3     & \cellcolor[rgb]{0.851, 0.851, 0.851}\textbf{6.725e+0}$^\ddagger$ & 6.622e+0$^\dagger$ & 5.044e+0$^\dagger$ & 6.611e+0$^\dagger$ & 5.592e+0$^\dagger$ & 6.523e+0$^\dagger$ & 6.628e+0$^\dagger$ & 6.563e+0$^\dagger$ & 6.708e+0$^\ddagger$ & 6.689e+0 \\
		& 5     & 1.773e+1 & 1.356e+1$^\dagger$ & 1.052e+1$^\dagger$ & 1.666e+1$^\dagger$ & 8.986e+0$^\dagger$ & 1.371e+1$^\dagger$ & 1.626e+1$^\dagger$ & 1.522e+1$^\dagger$ & 1.759e+1$^\dagger$ & \cellcolor[rgb]{0.851, 0.851, 0.851}\textbf{1.774e+1} \\
		DTLZ2$^{-1}$ & 8     & 4.509e+1$^\dagger$ & 2.981e+1$^\dagger$ & 1.805e+1$^\dagger$ & 4.212e+1$^\dagger$ & 2.078e+1$^\dagger$ & 2.285e+1$^\dagger$ & 2.409e+1$^\dagger$ & 1.858e+1$^\dagger$ & 4.569e+1$^\dagger$ & \cellcolor[rgb]{0.851, 0.851, 0.851}\textbf{4.941e+1} \\
		& 10    & 7.820e+1$^\dagger$ & 3.308e+1$^\dagger$ & 2.058e+1$^\dagger$ & 7.809e+1$^\dagger$ & 2.938e+1$^\dagger$ & 4.566e+1$^\dagger$ & 3.705e+1$^\dagger$ & 1.751e+1$^\dagger$ & 8.405e+1$^\dagger$ & \cellcolor[rgb]{0.851, 0.851, 0.851}\textbf{9.045e+1} \\
		& 15    & 9.507e+1$^\dagger$ & 4.900e+1$^\dagger$ & 2.131e+1$^\dagger$ & 1.408e+2$^\dagger$ & 5.276e+1$^\dagger$ & 5.544e+1$^\dagger$ & 3.028e+1$^\dagger$ & 2.045e+1$^\dagger$ & 8.979e+1$^\dagger$ & \cellcolor[rgb]{0.851, 0.851, 0.851}\textbf{1.491e+2} \\
		\hline
		& 3     & \cellcolor[rgb]{0.851, 0.851, 0.851}\textbf{6.359e+0}$^\ddagger$ & 6.266e+0$^\dagger$ & 5.005e+0$^\dagger$ & 6.251e+0$^\dagger$ & 5.918e+0$^\dagger$ & 6.139e+0$^\dagger$ & 6.297e+0$^\dagger$ & 6.234e+0$^\dagger$ & 6.340e+0$^\ddagger$ & 6.328e+0 \\
		& 5     & 1.636e+1 & 1.261e+1$^\dagger$ & 1.024e+1$^\dagger$ & 1.518e+1$^\dagger$ & 1.055e+1$^\dagger$ & 1.171e+1$^\dagger$ & 1.438e+1$^\dagger$ & 1.385e+1$^\dagger$ & 1.613e+1$^\dagger$ & \cellcolor[rgb]{0.851, 0.851, 0.851}\textbf{1.636e+1} \\
		DTLZ3$^{-1}$ & 8     & 4.095e+1$^\dagger$ & 2.725e+1$^\dagger$ & 1.674e+1$^\dagger$ & 3.732e+1$^\dagger$ & 2.353e+1$^\dagger$ & 1.232e+1$^\dagger$ & 2.075e+1$^\dagger$ & 1.501e+1$^\dagger$ & 4.009e+1$^\dagger$ & \cellcolor[rgb]{0.851, 0.851, 0.851}\textbf{4.415e+1} \\
		& 10    & 7.043e+1$^\dagger$ & 2.844e+1$^\dagger$ & 2.262e+1$^\dagger$ & 6.802e+1$^\dagger$ & 3.698e+1$^\dagger$ & 2.186e+1$^\dagger$ & 3.086e+1$^\dagger$ & 1.386e+1$^\dagger$ & 7.189e+1$^\dagger$ & \cellcolor[rgb]{0.851, 0.851, 0.851}\textbf{8.018e+1} \\
		& 15    & 8.678e+1$^\dagger$ & 4.581e+1$^\dagger$ & 2.245e+1$^\dagger$ & 1.248e+2$^\dagger$ & 3.980e+1$^\dagger$ & 3.594e+1$^\dagger$ & 3.176e+1$^\dagger$ & 1.778e+1$^\dagger$ & 8.691e+1$^\dagger$ & \cellcolor[rgb]{0.851, 0.851, 0.851}\textbf{1.313e+2} \\
		\hline
		& 3     & 6.578e+0$^\dagger$ & 6.623e+0$^\dagger$ & 4.255e+0$^\dagger$ & 6.632e+0$^\dagger$ & 5.316e+0$^\dagger$ & 6.538e+0$^\dagger$ & 6.669e+0$^\dagger$ & 6.621e+0$^\dagger$ & \cellcolor[rgb]{0.851, 0.851, 0.851}\textbf{6.706e+0}$^\ddagger$ & 6.694e+0 \\
		& 5     & 1.742e+1$^\dagger$ & 1.355e+1$^\dagger$ & 8.304e+0$^\dagger$ & 1.680e+1$^\dagger$ & 8.840e+0$^\dagger$ & 1.397e+1$^\dagger$ & 1.651e+1$^\dagger$ & 1.503e+1$^\dagger$ & 1.761e+1$^\dagger$ & \cellcolor[rgb]{0.851, 0.851, 0.851}\textbf{1.777e+1} \\
		DTLZ4$^{-1}$ & 8     & 4.339e+1$^\dagger$ & 3.225e+1$^\dagger$ & 1.218e+1$^\dagger$ & 4.255e+1$^\dagger$ & 1.675e+1$^\dagger$ & 2.280e+1$^\dagger$ & 2.872e+1$^\dagger$ & 1.197e+1$^\dagger$ & 4.578e+1$^\dagger$ & \cellcolor[rgb]{0.851, 0.851, 0.851}\textbf{4.856e+1} \\
		& 10    & 7.816e+1$^\dagger$ & 5.378e+1$^\dagger$ & 1.911e+1$^\dagger$ & 7.831e+1$^\dagger$ & 1.995e+1$^\dagger$ & 4.790e+1$^\dagger$ & 4.238e+1$^\dagger$ & 1.472e+1$^\dagger$ & 8.423e+1$^\dagger$ & \cellcolor[rgb]{0.851, 0.851, 0.851}\textbf{8.781e+1} \\
		& 15    & 8.953e+1$^\dagger$ & 4.729e+1$^\dagger$ & 2.226e+1$^\dagger$ & 1.384e+2$^\dagger$ & 3.040e+1$^\dagger$ & 5.869e+1$^\dagger$ & 1.468e+1$^\dagger$ & 3.091e+1$^\dagger$ & 9.467e+1$^\dagger$ & \cellcolor[rgb]{0.851, 0.851, 0.851}\textbf{1.460e+2} \\
		\hline
	\end{tabular}%
	\begin{tablenotes}
		\item[1] According to Wilcoxon's rank sum test, $^\dagger$ and $^\ddagger$ indicates whether the corresponding algorithm is significantly worse or better than MOEA/AD respectively.
	\end{tablenotes}
\end{table*}

The comparison results on the DTLZ$^{-1}$ problem instances are given in~\pref{tab:iDTLZ_2}. Similar to the observations in~\pref{tab:DTLZ_2}, MOEA/AD is the best algorithm which wins on almost all comparisons (150 out of 153) except for the 3-objective DTLZ1$^{-1}$, DTLZ3$^{-1}$ and DTLZ4$^{-1}$ instances. In particular, MOEA/D-IPBI shows better performance than MOEA/AD on the 3-objective DTLZ2$^{-1}$ and DTLZ3$^{-1}$ instances, while Two$\_$Arch2 outperforms MOEA/AD on the 3-objective DTLZ2$^{-1}$ to DTLZ4$^{-1}$ instances. The inferior performance of MOEA/AD might be partially caused by the disturbance from its normalization procedure. As discussed in~\cite{LiDAY17}, uniformly sampled weight vectors upon the simplex tend to guide the population towards the boundaries of a hyperspherical PF (e.g., DTLZ2$^{-1}$ to DTLZ4$^{-1}$). This explains the relatively good performance obtained by Two$\_$Arch2 and VaEA which do not rely on a set of fixed weight vectors. However, we also notice that the performance of PICEA-g, HypE and KnEA are not satisfied under this setting. Although Global WASF-GA also uses the nadir point as the reference point in its scalarizing function like MOEA/AD, it fails to obtain comparable performance due to its poor diversity maintenance scheme.

\begin{table*}[!t]
	\renewcommand{\arraystretch}{0.8}
	\setlength\tabcolsep{5pt}
	\tiny
	\centering
	\caption{Comparison Results of MOEA/AD and 9 Peer Algorithms on WFG$^{-1}$ Problem Instances.}
	\label{tab:iWFG_2}
	\begin{tabular}{cccccccccccc}
		\hline
		Problem & m     & ~~IPBI~~ & GWASF & PICEA-g & VaEA  & ~HypE~  & KnEA  & NSGA-III & $\theta$-DEA & TwoArch2 & MOEA/AD \\
		\hline
		& 3     & 3.704e+0$^\dagger$ & \cellcolor[rgb]{0.851, 0.851, 0.851}\textbf{4.219e+0}$^\ddagger$ & 3.370e+0$^\dagger$ & 3.151e+0$^\dagger$ & 2.353e+0$^\dagger$ & 4.154e+0$^\ddagger$ & 3.041e+0$^\dagger$ & 3.007e+0$^\dagger$ & 3.886e+0 & 3.892e+0 \\
		& 5     & 5.402e+0$^\dagger$ & 6.713e+0 & 5.652e+0$^\dagger$ & 5.121e+0$^\dagger$ & 3.014e+0$^\dagger$ & 5.915e+0$^\dagger$ & 3.537e+0$^\dagger$ & 5.067e+0$^\dagger$ & 3.110e+0$^\dagger$ & \cellcolor[rgb]{0.851, 0.851, 0.851}\textbf{6.727e+0} \\
		WFG1$^{-1}$ & 8     & 7.430e+0$^\ddagger$ & 2.832e+0$^\dagger$ & 7.278e+0$^\ddagger$ & 6.675e+0$^\ddagger$ & 2.944e+0$^\dagger$ & \cellcolor[rgb]{0.851, 0.851, 0.851}\textbf{7.614e+0}$^\ddagger$ & 3.479e+0$^\dagger$ & 4.742e+0$^\ddagger$ & 2.972e+0$^\dagger$ & 3.920e+0 \\
		& 10    & 1.011e+1$^\ddagger$ & 3.481e+0$^\dagger$ & \cellcolor[rgb]{0.851, 0.851, 0.851}\textbf{1.085e+1}$^\ddagger$ & 8.997e+0$^\ddagger$ & 2.957e+0$^\dagger$ & 1.071e+1$^\ddagger$ & 3.602e+0 & 5.362e+0$^\ddagger$ & 2.946e+0$^\dagger$ & 3.735e+0 \\
		& 15    & \cellcolor[rgb]{0.851, 0.851, 0.851}\textbf{1.434e+1$^\ddagger$} & 1.849e+0$^\dagger$ & 1.091e+1$^\ddagger$ & 1.097e+1$^\ddagger$ & 4.132e+0$^\dagger$ & 8.360e+0$^\ddagger$ & 3.397e+0$^\dagger$ & 3.745e+0$^\dagger$ & 2.993e+0$^\dagger$ & 4.693e+0 \\
		\hline
		& 3     & 5.987e+0$^\dagger$ & 6.084e+0$^\dagger$ & 4.673e+0$^\dagger$ & 6.091e+0$^\dagger$ & 5.878e+0$^\dagger$ & 5.754e+0$^\dagger$ & 6.104e+0$^\dagger$ & 6.108e+0$^\dagger$ & 6.069e+0$^\dagger$ & \cellcolor[rgb]{0.851, 0.851, 0.851}\textbf{6.134e+0} \\
		& 5     & 9.557e+0$^\dagger$ & 1.016e+1$^\dagger$ & 5.210e+0$^\dagger$ & 1.099e+1$^\dagger$ & 8.914e+0$^\dagger$ & 8.839e+0$^\dagger$ & 9.498e+0$^\dagger$ & 1.031e+1$^\dagger$ & 8.272e+0$^\dagger$ & \cellcolor[rgb]{0.851, 0.851, 0.851}\textbf{1.117e+1} \\
		WFG2$^{-1}$ & 8     & 1.530e+1$^\dagger$ & 6.643e+0$^\dagger$ & 4.397e+0$^\dagger$ & \cellcolor[rgb]{0.851, 0.851, 0.851}\textbf{1.799e+1}$^\dagger$ & 9.955e+0$^\dagger$ & 1.125e+1$^\dagger$ & 8.135e+0$^\dagger$ & 9.735e+0$^\dagger$ & 7.879e+0$^\dagger$ & 1.746e+1 \\
		& 10    & 1.908e+1$^\dagger$ & 7.388e+0$^\dagger$ & 4.519e+0$^\dagger$ & 2.390e+1$^\dagger$ & 1.241e+1$^\dagger$ & 1.569e+1$^\dagger$ & 8.431e+0$^\dagger$ & 9.237e+0$^\dagger$ & 7.647e+0$^\dagger$ & \cellcolor[rgb]{0.851, 0.851, 0.851}\textbf{2.447e+1} \\
		& 15    & 2.794e+1$^\dagger$ & 3.472e+0$^\dagger$ & 3.366e+0$^\dagger$ & 3.124e+1$^\dagger$ & 1.760e+1$^\dagger$ & 1.374e+1$^\dagger$ & 8.035e+0$^\dagger$ & 6.240e+0$^\dagger$ & 7.501e+0$^\dagger$ & \cellcolor[rgb]{0.851, 0.851, 0.851}\textbf{3.587e+1} \\
		\hline
		& 3     & 4.779e+0$^\dagger$ & 5.445e+0 & 3.695e+0$^\dagger$ & 5.370e+0$^\dagger$ & 4.373e+0$^\dagger$ & 4.430e+0$^\dagger$ & 5.283e+0$^\dagger$ & 5.327e+0$^\dagger$ & 5.447e+0 & \cellcolor[rgb]{0.851, 0.851, 0.851}\textbf{5.447e+0} \\
		& 5     & 7.877e+0$^\dagger$ & 1.043e+1$^\dagger$ & 4.872e+0$^\dagger$ & 1.051e+1$^\dagger$ & 6.368e+0$^\dagger$ & 7.286e+0$^\dagger$ & 8.213e+0$^\dagger$ & 7.307e+0$^\dagger$ & 1.005e+1$^\dagger$ & \cellcolor[rgb]{0.851, 0.851, 0.851}\textbf{1.104e+1} \\
		WFG3$^{-1}$ & 8     & 1.124e+1$^\dagger$ & 8.709e+0$^\dagger$ & 4.670e+0$^\dagger$ & 1.959e+1$^\dagger$ & 7.962e+0$^\dagger$ & 1.027e+1$^\dagger$ & 7.321e+0$^\dagger$ & 3.576e+0$^\dagger$ & 1.144e+1$^\dagger$ & \cellcolor[rgb]{0.851, 0.851, 0.851}\textbf{1.986e+1} \\
		& 10    & 1.469e+1$^\dagger$ & 1.076e+1$^\dagger$ & 5.122e+0$^\dagger$ & 2.899e+1$^\dagger$ & 1.027e+1$^\dagger$ & 1.597e+1$^\dagger$ & 8.023e+0$^\dagger$ & 2.679e+0$^\dagger$ & 1.313e+1$^\dagger$ & \cellcolor[rgb]{0.851, 0.851, 0.851}\textbf{2.983e+1} \\
		& 15    & 1.960e+1$^\dagger$ & 5.816e+0$^\dagger$ & 3.983e+0$^\dagger$ & 3.957e+1$^\dagger$ & 1.232e+1$^\dagger$ & 1.640e+1$^\dagger$ & 6.675e+0$^\dagger$ & 8.462e-1$^\dagger$ & 1.963e+1$^\dagger$ & \cellcolor[rgb]{0.851, 0.851, 0.851}\textbf{4.094e+1} \\
		\hline
		& 3     & 6.694e+0$^\ddagger$ & 6.616e+0$^\dagger$ & 5.692e+0$^\dagger$ & 6.474e+0$^\dagger$ & 5.746e+0$^\dagger$ & 6.016e+0$^\dagger$ & 6.390e+0$^\dagger$ & 6.466e+0$^\dagger$ & \cellcolor[rgb]{0.851, 0.851, 0.851}\textbf{6.705e+0}$^\ddagger$ & 6.669e+0 \\
		& 5     & 1.723e+1$^\dagger$ & 1.362e+1$^\dagger$ & 1.392e+1$^\dagger$ & 1.592e+1$^\dagger$ & 1.081e+1$^\dagger$ & 8.779e+0$^\dagger$ & 1.477e+1$^\dagger$ & 1.500e+1$^\dagger$ & \cellcolor[rgb]{0.851, 0.851, 0.851}\textbf{1.761e+1} & 1.760e+1 \\
		WFG4$^{-1}$ & 8     & 4.044e+1$^\dagger$ & 2.959e+1$^\dagger$ & 2.182e+1$^\dagger$ & 4.152e+1$^\dagger$ & 1.669e+1$^\dagger$ & 1.542e+1$^\dagger$ & 2.158e+1$^\dagger$ & 1.943e+1$^\dagger$ & 4.518e+1$^\dagger$ & \cellcolor[rgb]{0.851, 0.851, 0.851}\textbf{4.687e+1} \\
		& 10    & 6.781e+1$^\dagger$ & 3.521e+1$^\dagger$ & 3.171e+1$^\dagger$ & 7.704e+1$^\dagger$ & 2.403e+1$^\dagger$ & 2.429e+1$^\dagger$ & 3.532e+1$^\dagger$ & 1.417e+1$^\dagger$ & 7.737e+1$^\dagger$ & \cellcolor[rgb]{0.851, 0.851, 0.851}\textbf{8.534e+1} \\
		& 15    & 8.277e+1$^\dagger$ & 5.149e+1$^\dagger$ & 1.333e+1$^\dagger$ & 1.444e+2$^\dagger$ & 3.998e+1$^\dagger$ & 2.686e+1$^\dagger$ & 3.787e+1$^\dagger$ & 3.637e+0$^\dagger$ & 5.919e+1$^\dagger$ & \cellcolor[rgb]{0.851, 0.851, 0.851}\textbf{1.486e+2} \\
		\hline
		& 3     & 6.683e+0$^\ddagger$ & 6.605e+0$^\dagger$ & 5.708e+0$^\dagger$ & 6.487e+0$^\dagger$ & 5.414e+0$^\dagger$ & 6.105e+0$^\dagger$ & 6.444e+0$^\dagger$ & 6.515e+0$^\dagger$ & \cellcolor[rgb]{0.851, 0.851, 0.851}\textbf{6.686e+0}$^\ddagger$ & 6.665e+0 \\
		& 5     & 1.723e+1$^\dagger$ & 1.373e+1$^\dagger$ & 1.313e+1$^\dagger$ & 1.648e+1$^\dagger$ & 1.075e+1$^\dagger$ & 8.647e+0$^\dagger$ & 1.554e+1$^\dagger$ & 1.496e+1$^\dagger$ & 1.737e+1$^\dagger$ & \cellcolor[rgb]{0.851, 0.851, 0.851}\textbf{1.769e+1} \\
		WFG5$^{-1}$ & 8     & 4.065e+1$^\dagger$ & 3.111e+1$^\dagger$ & 2.486e+1$^\dagger$ & 4.026e+1$^\dagger$ & 1.852e+1$^\dagger$ & 1.564e+1$^\dagger$ & 2.444e+1$^\dagger$ & 1.768e+1$^\dagger$ & 4.329e+1$^\dagger$ & \cellcolor[rgb]{0.851, 0.851, 0.851}\textbf{4.771e+1} \\
		& 10    & 6.840e+1$^\dagger$ & 3.004e+1$^\dagger$ & 3.780e+1$^\dagger$ & 7.694e+1$^\dagger$ & 2.641e+1$^\dagger$ & 2.285e+1$^\dagger$ & 3.573e+1$^\dagger$ & 1.111e+1$^\dagger$ & 7.438e+1$^\dagger$ & \cellcolor[rgb]{0.851, 0.851, 0.851}\textbf{8.627e+1} \\
		& 15    & 8.410e+1$^\dagger$ & 4.579e+1$^\dagger$ & 6.327e+1$^\dagger$ & 1.412e+2$^\dagger$ & 4.360e+1$^\dagger$ & 2.790e+1$^\dagger$ & 3.888e+1$^\dagger$ & 5.382e+0$^\dagger$ & 6.110e+1$^\dagger$ & \cellcolor[rgb]{0.851, 0.851, 0.851}\textbf{1.460e+2} \\
		\hline
		& 3     & 6.693e+0$^\ddagger$ & 6.619e+0$^\dagger$ & 5.775e+0$^\dagger$ & 6.564e+0$^\dagger$ & 5.110e+0$^\dagger$ & 6.317e+0$^\dagger$ & 6.539e+0$^\dagger$ & 6.548e+0$^\dagger$ & \cellcolor[rgb]{0.851, 0.851, 0.851}\textbf{6.703e+0}$^\ddagger$ & 6.676e+0 \\
		& 5     & 1.723e+1$^\dagger$ & 1.357e+1$^\dagger$ & 1.372e+1$^\dagger$ & 1.681e+1$^\dagger$ & 9.565e+0$^\dagger$ & 9.297e+0$^\dagger$ & 1.589e+1$^\dagger$ & 1.506e+1$^\dagger$ & 1.753e+1$^\dagger$ & \cellcolor[rgb]{0.851, 0.851, 0.851}\textbf{1.771e+1} \\
		WFG6$^{-1}$ & 8     & 4.049e+1$^\dagger$ & 3.076e+1$^\dagger$ & 2.783e+1$^\dagger$ & 3.954e+1$^\dagger$ & 1.706e+1$^\dagger$ & 1.561e+1$^\dagger$ & 2.435e+1$^\dagger$ & 1.628e+1$^\dagger$ & 4.460e+1$^\dagger$ & \cellcolor[rgb]{0.851, 0.851, 0.851}\textbf{4.810e+1} \\
		& 10    & 6.792e+1$^\dagger$ & 3.684e+1$^\dagger$ & 4.418e+1$^\dagger$ & 7.574e+1$^\dagger$ & 2.455e+1$^\dagger$ & 2.014e+1$^\dagger$ & 3.586e+1$^\dagger$ & 9.602e+0$^\dagger$ & 7.944e+1$^\dagger$ & \cellcolor[rgb]{0.851, 0.851, 0.851}\textbf{8.714e+1} \\
		& 15    & 8.136e+1$^\dagger$ & 4.105e+1$^\dagger$ & 5.323e+1$^\dagger$ & 1.383e+2$^\dagger$ & 3.667e+1$^\dagger$ & 2.072e+1$^\dagger$ & 4.102e+1$^\dagger$ & 6.963e+0$^\dagger$ & 8.900e+1$^\dagger$ & \cellcolor[rgb]{0.851, 0.851, 0.851}\textbf{1.460e+2} \\
		\hline
		& 3     & 6.694e+0$^\ddagger$ & 6.610e+0$^\dagger$ & 5.965e+0$^\dagger$ & 6.519e+0$^\dagger$ & 5.814e+0$^\dagger$ & 5.748e+0$^\dagger$ & 6.455e+0$^\dagger$ & 6.537e+0$^\dagger$ & \cellcolor[rgb]{0.851, 0.851, 0.851}\textbf{6.706e+0}$^\ddagger$ & 6.665e+0 \\
		& 5     & 1.723e+1$^\dagger$ & 1.357e+1$^\dagger$ & 1.418e+1$^\dagger$ & 1.648e+1$^\dagger$ & 1.000e+1$^\dagger$ & 8.338e+0$^\dagger$ & 1.537e+1$^\dagger$ & 1.501e+1$^\dagger$ & 1.755e+1$^\dagger$ & \cellcolor[rgb]{0.851, 0.851, 0.851}\textbf{1.765e+1} \\
		WFG7$^{-1}$ & 8     & 4.046e+1$^\dagger$ & 3.132e+1$^\dagger$ & 2.392e+1$^\dagger$ & 3.983e+1$^\dagger$ & 1.658e+1$^\dagger$ & 1.492e+1$^\dagger$ & 2.272e+1$^\dagger$ & 1.763e+1$^\dagger$ & 4.395e+1$^\dagger$ & \cellcolor[rgb]{0.851, 0.851, 0.851}\textbf{4.690e+1} \\
		& 10    & 6.787e+1$^\dagger$ & 4.803e+1$^\dagger$ & 3.421e+1$^\dagger$ & 7.558e+1$^\dagger$ & 2.308e+1$^\dagger$ & 2.199e+1$^\dagger$ & 3.361e+1$^\dagger$ & 1.236e+1$^\dagger$ & 7.549e+1$^\dagger$ & \cellcolor[rgb]{0.851, 0.851, 0.851}\textbf{8.515e+1} \\
		& 15    & 8.381e+1$^\dagger$ & 3.392e+1$^\dagger$ & 2.380e+1$^\dagger$ & 1.408e+2$^\dagger$ & 4.449e+1$^\dagger$ & 1.739e+1$^\dagger$ & 3.384e+1$^\dagger$ & 4.568e+0$^\dagger$ & 8.796e+1$^\dagger$ & \cellcolor[rgb]{0.851, 0.851, 0.851}\textbf{1.475e+2} \\
		\hline
		& 3     & \cellcolor[rgb]{0.851, 0.851, 0.851}\textbf{6.693e+0}$^\ddagger$ & 6.619e+0$^\dagger$ & 6.073e+0$^\dagger$ & 6.554e+0$^\dagger$ & 6.048e+0$^\dagger$ & 6.449e+0$^\dagger$ & 6.566e+0$^\dagger$ & 6.550e+0$^\dagger$ & 6.690e+0$^\ddagger$ & 6.683e+0 \\
		& 5     & 1.722e+1$^\dagger$ & 1.355e+1$^\dagger$ & 1.461e+1$^\dagger$ & 1.700e+1$^\dagger$ & 9.001e+0$^\dagger$ & 1.197e+1$^\dagger$ & 1.614e+1$^\dagger$ & 1.485e+1$^\dagger$ & 1.749e+1$^\dagger$ & \cellcolor[rgb]{0.851, 0.851, 0.851}\textbf{1.774e+1} \\
		WFG8$^{-1}$ & 8     & 4.054e+1$^\dagger$ & 3.160e+1$^\dagger$ & 3.148e+1$^\dagger$ & 4.329e+1$^\dagger$ & 1.759e+1$^\dagger$ & 2.354e+1$^\dagger$ & 2.548e+1$^\dagger$ & 1.746e+1$^\dagger$ & 4.521e+1$^\dagger$ & \cellcolor[rgb]{0.851, 0.851, 0.851}\textbf{4.799e+1} \\
		& 10    & 6.805e+1$^\dagger$ & 3.128e+1$^\dagger$ & 5.119e+1$^\dagger$ & 7.992e+1$^\dagger$ & 2.518e+1$^\dagger$ & 5.337e+1$^\dagger$ & 3.799e+1$^\dagger$ & 1.205e+1$^\dagger$ & 8.332e+1$^\dagger$ & \cellcolor[rgb]{0.851, 0.851, 0.851}\textbf{8.708e+1} \\
		& 15    & 8.302e+1$^\dagger$ & 4.373e+1$^\dagger$ & 5.093e+1$^\dagger$ & \cellcolor[rgb]{0.851, 0.851, 0.851}\textbf{1.529e+2}$^\ddagger$ & 3.877e+1$^\dagger$ & 6.124e+1$^\dagger$ & 4.135e+1$^\dagger$ & 6.750e+0$^\dagger$ & 1.275e+2$^\dagger$ & 1.464e+2 \\
		\hline
		& 3     & 6.665e+0$^\ddagger$ & 6.565e+0$^\dagger$ & 6.332e+0$^\dagger$ & 6.520e+0$^\dagger$ & 5.345e+0$^\dagger$ & 6.305e+0$^\dagger$ & 6.498e+0$^\dagger$ & 6.529e+0$^\dagger$ & \cellcolor[rgb]{0.851, 0.851, 0.851}\textbf{6.675e+0}$^\ddagger$ & 6.634e+0 \\
		& 5     & 1.720e+1$^\dagger$ & 1.382e+1$^\dagger$ & 1.535e+1$^\dagger$ & 1.673e+1$^\dagger$ & 1.089e+1$^\dagger$ & 1.429e+1$^\dagger$ & 1.597e+1$^\dagger$ & 1.509e+1$^\dagger$ & 1.741e+1$^\dagger$ & \cellcolor[rgb]{0.851, 0.851, 0.851}\textbf{1.748e+1} \\
		WFG9$^{-1}$ & 8     & 4.060e+1$^\dagger$ & 3.124e+1$^\dagger$ & 3.145e+1$^\dagger$ & 4.230e+1$^\dagger$ & 1.971e+1$^\dagger$ & 1.920e+1$^\dagger$ & 2.644e+1$^\dagger$ & 1.322e+1$^\dagger$ & 4.404e+1$^\dagger$ & \cellcolor[rgb]{0.851, 0.851, 0.851}\textbf{4.692e+1} \\
		& 10    & 6.905e+1$^\dagger$ & 4.819e+1$^\dagger$ & 5.044e+1$^\dagger$ & 7.767e+1$^\dagger$ & 2.453e+1$^\dagger$ & 3.322e+1$^\dagger$ & 4.016e+1$^\dagger$ & 8.968e+0$^\dagger$ & 7.848e+1$^\dagger$ & \cellcolor[rgb]{0.851, 0.851, 0.851}\textbf{8.549e+1} \\
		& 15    & 8.989e+1$^\dagger$ & 4.867e+1$^\dagger$ & 6.333e+1$^\dagger$ & 1.432e+2 & 4.166e+1$^\dagger$ & 4.476e+1$^\dagger$ & 5.153e+1$^\dagger$ & 7.383e+0$^\dagger$ & 1.175e+2$^\dagger$ & \cellcolor[rgb]{0.851, 0.851, 0.851}\textbf{1.444e+2} \\
		\hline
	\end{tabular}%
	\begin{tablenotes}
		\item[1] According to Wilcoxon's rank sum test, $^\dagger$ and $^\ddagger$ indicates whether the corresponding algorithm is significantly worse or better than MOEA/AD respectively.
	\end{tablenotes}
\end{table*}

\begin{figure*}[!t]
	\centering
	\subfloat[]{\includegraphics[width=.5\linewidth]{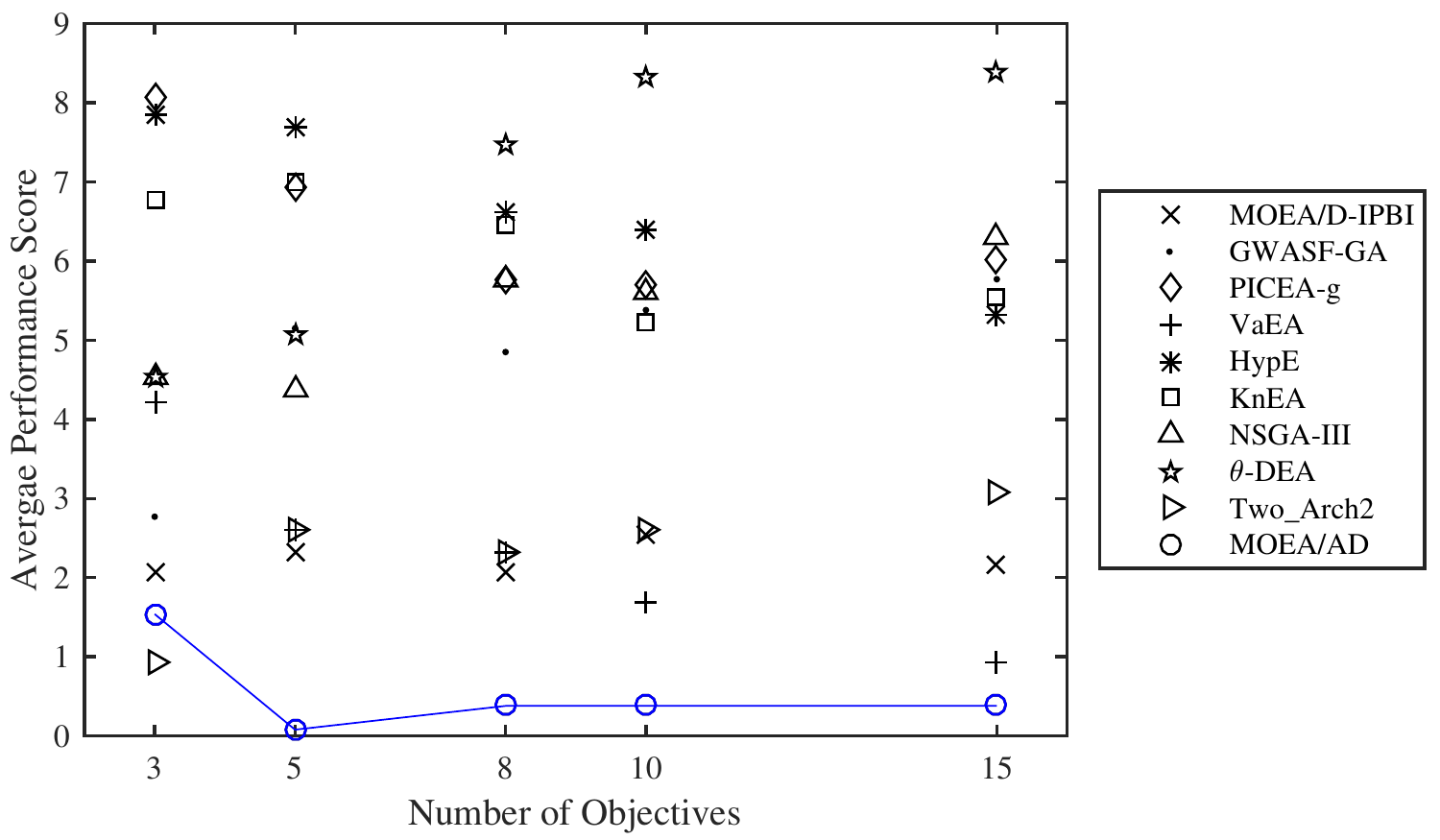}}
	\subfloat[]{\includegraphics[width=.5\linewidth]{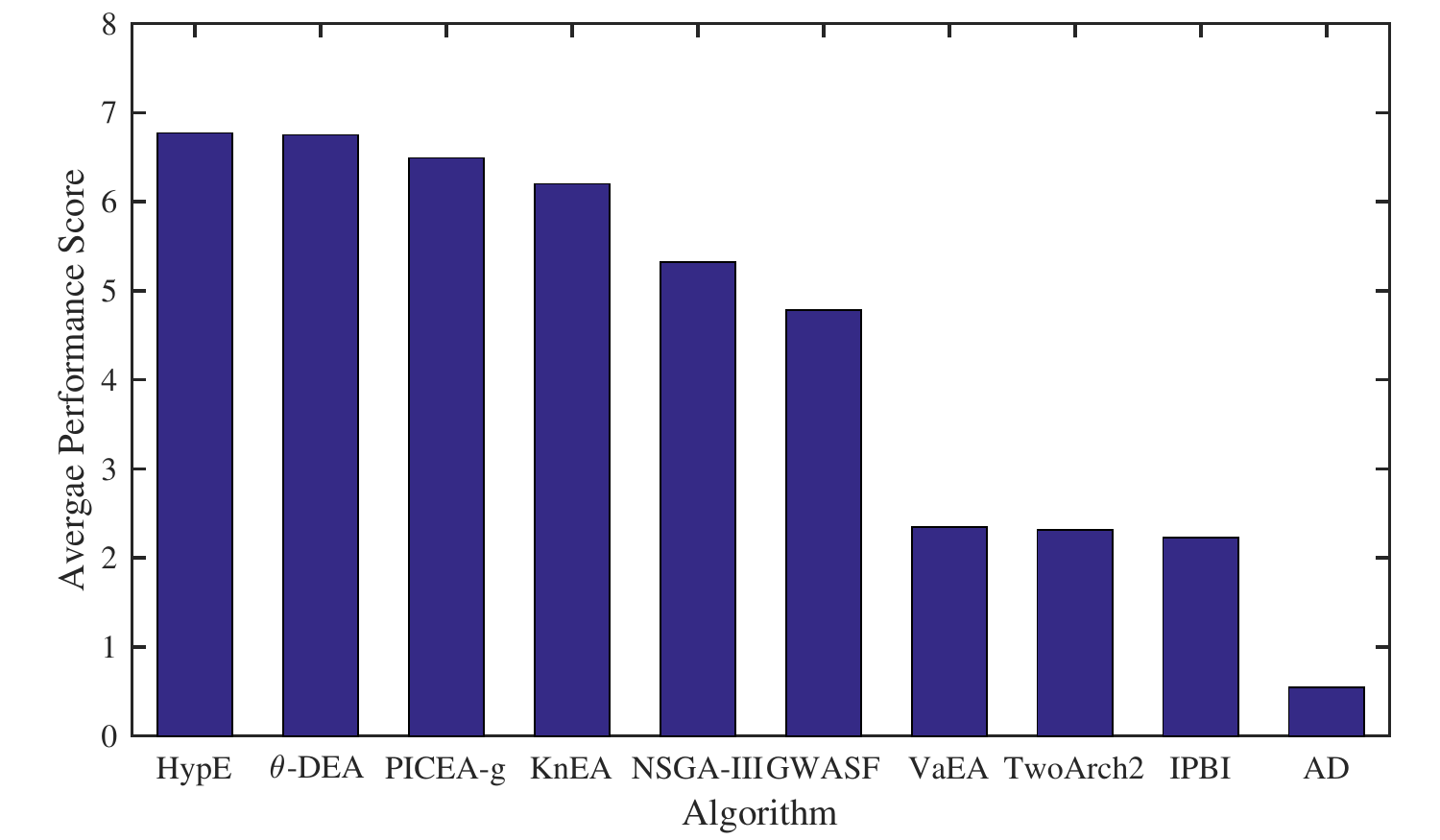}}
	\caption{(a) Average performance scores on different number of objectives over DTLZ$^{-1}$ and WFG$^{-1}$ problem instances. (b) Average performance scores over all DTLZ$^{-1}$ and WFG$^{-1}$ problem instances.}
	\label{fig:pscore_i}
\end{figure*}

The HV results on WFG$^{-1}$ problem instances are displayed in \pref{tab:iWFG_2}. From \pref{tab:iWFG_2}, MOEA/AD achieves the best overall performance on the WFG$^{-1}$ test suite, where it significantly out performs its competitor in 369 out of 450 comparisons. WFG1$^{-1}$ and WFG2$^{-1}$ have quite complex PF shapes. The best algorithms on WFG1$^{-1}$ differ with the number of objectives. Global WASF-GA and MOEA/AD show the best HV results on 3- and 5-objective WFG1$^{-1}$ instances respectively. When the objectives are more than 5, MOEA/D-IPBI, PICEA-g, VaEA and KnEA become the best four algorithms, three of which do not use fixed weight vectors to guide the search. In contrast, MOEA/AD is the best algorithm on all WFG2$^{-1}$ instances. Even though VaEA has a slightly larger mean HV metric value on 8-objective WFG2$^{-1}$, MOEA/AD is shown to be significantly better in all 45 comparisons according to Wilcoxon's rank sum test. The PFs of WFG3$^{-1}$ instances are hyperplanes, which are perfect for decomposition-based algorithms using the nadir point. MOEA/AD remains being the best algorithm on all 5 problem instances, while MOEA/D-IPBI does not perform as good as expected. It is also worth noting that Two$\_$Arch2 never beats MOEA/AD on WFG1$^{-1}$ to WFG3$^{-1}$ instances. The PFs of WFG4$^{-1}$ to WFG9$^{-1}$ are hyperspheres centered at the nadir point. MOEA/AD is significantly better than the other algorithms on all 5- to 15-objective WFG4$^{-1}$ to WFG9$^{-1}$ instances except that it is outperformed by Two$\_$Arch2 and VaEA on 5-objective WFG4$^{-1}$ and 15-objective WFG8$^{-1}$ respectively. The reason why MOEA/AD outperforms the other algorithms, including Two$\_$Arch2, is not simply due to the co-evolving populations but mainly because of the adversarial search directions and the well constructed collaboration for reproduction. Similar to the situation on DTLZ$^{-1}$ test suite, Two$\_$Arch2 and MOEA/D-IPBI are the best two algorithms on 3-objective WFG4$^{-1}$ to WFG9$^{-1}$ instances. 

We calculate the average performance scores of different algorithms on DTLZ$^{-1}$ and WFG$^{-1}$ test suites and display the results in \pref{fig:pscore_i}. As shown in \pref{fig:pscore_i}(b), MOEA/AD, whose average performance score is four times smaller than the runner-up, remains the best among all test algorithms. It is worth noting that the final ranking of MOEA/D and Global WASF-GA increase dramatically compared with \pref{fig:pscore} due to the use of nadir point. However, the ranking of $\theta$-DEA, which adopts fixed weight vectors starting from the ideal point, drops significantly. More specifically, from \pref{fig:pscore_i}(a), MOEA/AD achieves the best average performance scores on problems with more than three objectives and obtains the second best results problems with 3 and 15 objectives. Following MOEA/AD, the performance of Two$\_$Arch2 degenerates when the number of objectives increases, whereas, the performance of VaEA is improved with the number of objectives. 

\subsection{Performance Comparisons with Three Variants}

\begin{algorithm}[!t]
	\caption{$\mathsf{MatingSelectionV2}(S_c,S_d,i,M,R,C,W,B)$}
	\label{alg:matingSelect_v2}
	$pop\leftarrow\mathsf{PopSelection}(S_c,S_d,i,M,W)$;\\
	\uIf{$rand<\delta$}{
		$S_p\leftarrow\emptyset$;\\
		\uIf{$pop==1$}{
			\For{$j\leftarrow 1$ \KwTo $T$}{
				$S_p\leftarrow S_p\cup \{\mathbf{x}_d^{B[i][j]}\}$;\\
			}
			$\mathbf{x}^r\leftarrow$ Randomly select a solution from $S_p$;\\
			$\overline{S}\leftarrow\{\mathbf{x}_d^i,\mathbf{x}^r\}$;\\
		}
		\Else{
			\For{$j\leftarrow 1$ \KwTo $T$}{
				$S_p\leftarrow S_p\cup \{\mathbf{x}_c^{B[M[i]][j]}\}$;\\
			}
			$\mathbf{x}^r\leftarrow$ Randomly select a solution from $S_p$;\\
			$\overline{S}\leftarrow\{\mathbf{x}_c^{M[i]},\mathbf{x}^r$\};\\	
		}
	}
	\Else{
		\uIf{$pop==1$}{
			$\mathbf{x}^r\leftarrow$ Randomly select a solution from $S_d$;\\
			$\overline{S}\leftarrow\{\mathbf{x}_d^i,\mathbf{x}^r$\};\\
		}
		\Else{
			$\mathbf{x}^r\leftarrow$ Randomly select a solution from $S_c$;\\
			$\overline{S}\leftarrow\{\mathbf{x}_c^{M[i]},\mathbf{x}^r$\};\\	
		}
	}
	\Return $\overline{S}$
\end{algorithm}

In our proposed MOEA/AD, there are two major aspects that contribute to the complementary effect of two co-evolving populations: one is the use of two scalarizing functions which results in the adversarial search directions; the other is the sophisticated mating selection process. The effectiveness of the prior aspect has been validated in the comparisons with the MOEA/D variants with a single scalarizing function. To further investigate the effectiveness of the sophisticated mating selection process, we design the following three variants to validate its three major components, i.e., the pairing step, the mating selection step and the principal parent selection. 
\begin{itemize}
	\item MOEA/AD-$v1$: it replaces the two-level stable matching in line 13 of \pref{alg:moead-2p} with random matching. In particular, $M$ is set as a random permutation among 1 to $N$ and $R[i]=1$ for all $i\in\{1,\cdots,N\}$.
	\item MOEA/AD-$v2$: it does not consider the collaboration between $S_d$ and $S_c$ in the mating selection step. In particular, it randomly selects the mating parents from $S_d$ and $S_c$ according to the principal parent. The pseudo code is given in~\pref{alg:matingSelect_v2}.
	\item MOEA/AD-$v3$: it only considers one criteria, i.e., subproblem's relative improvement, to select the principal parent solution. In short, line 6-11 of \pref{alg:popSelect} are replaced by $pop\leftarrow$ Randomly select from $\{1,2\}$.
\end{itemize} 

\begin{table}[!t]
	\renewcommand{\arraystretch}{0.8}
	\setlength\tabcolsep{5pt}
	\centering
	\caption{Comparison Results of MOEA/AD and its Three Variants on DTLZ Problem Instances.}
	\label{tab:DTLZ_v}
	\begin{tabular}{cccccc}
		\hline
		Problem & m     & AD-$v1$ & AD-$v2$ & AD-$v3$ & AD \\
		\hline
		& 3     & 7.785e+0$^\dagger$ & 7.785e+0$^\dagger$ & 7.777e+0$^\dagger$ & \cellcolor[rgb]{0.851, 0.851, 0.851}\textbf{7.787e+0} \\
		& 5     & 3.197e+1$^\dagger$ & 3.196e+1$^\dagger$ & 3.191e+1$^\dagger$ & \cellcolor[rgb]{0.851, 0.851, 0.851}\textbf{3.197e+1} \\
		DTLZ1 & 8     & 2.560e+2$^\dagger$ & 2.491e+2$^\dagger$ & 2.499e+2$^\dagger$ & \cellcolor[rgb]{0.851, 0.851, 0.851}\textbf{2.560e+2} \\
		& 10    & 1.008e+3$^\dagger$ & 9.767e+2$^\dagger$ & 1.006e+3$^\dagger$ & \cellcolor[rgb]{0.851, 0.851, 0.851}\textbf{1.024e+3} \\
		& 15    & 3.055e+4$^\dagger$ & 3.186e+4$^\dagger$ & 3.098e+4$^\dagger$ & \cellcolor[rgb]{0.851, 0.851, 0.851}\textbf{3.270e+4} \\
		\hline
		& 3     & 7.412e+0 & 7.412e+0 & 7.411e+0$^\dagger$ & \cellcolor[rgb]{0.851, 0.851, 0.851}\textbf{7.412e+0} \\
		& 5     & 3.170e+1 & \cellcolor[rgb]{0.851, 0.851, 0.851}\textbf{3.170e+1} & 3.170e+1$^\dagger$ & 3.170e+1 \\
		DTLZ2 & 8     & 2.558e+2$^\dagger$ & 2.558e+2$^\dagger$ & 2.558e+2$^\dagger$ & \cellcolor[rgb]{0.851, 0.851, 0.851}\textbf{2.558e+2} \\
		& 10    & 1.024e+3$^\dagger$ & 1.024e+3$^\dagger$ & 1.024e+3 & \cellcolor[rgb]{0.851, 0.851, 0.851}\textbf{1.024e+3} \\
		& 15    & \cellcolor[rgb]{0.851, 0.851, 0.851}\textbf{3.276e+4} & 3.276e+4 & 3.276e+4 & 3.276e+4 \\
		\hline
		& 3     & \cellcolor[rgb]{0.851, 0.851, 0.851}\textbf{7.405e+0} & 7.239e+0 & 7.405e+0 & 7.403e+0 \\
		& 5     & 3.097e+1$^\dagger$ & 2.877e+1$^\dagger$ & 3.092e+1$^\dagger$ & \cellcolor[rgb]{0.851, 0.851, 0.851}\textbf{3.169e+1} \\
		DTLZ3 & 8     & 2.125e+2$^\dagger$ & 2.330e+2$^\dagger$ & 2.058e+2$^\dagger$ & \cellcolor[rgb]{0.851, 0.851, 0.851}\textbf{2.558e+2} \\
		& 10    & 8.783e+2$^\dagger$ & 8.356e+2$^\dagger$ & 6.739e+2$^\dagger$ & \cellcolor[rgb]{0.851, 0.851, 0.851}\textbf{1.024e+3} \\
		& 15    & 2.761e+4$^\dagger$ & 2.574e+4$^\dagger$ & 2.681e+4$^\dagger$ & \cellcolor[rgb]{0.851, 0.851, 0.851}\textbf{3.276e+4} \\
		\hline
		& 3     & 7.410e+0$^\dagger$ & 7.411e+0 & 7.410e+0$^\dagger$ & \cellcolor[rgb]{0.851, 0.851, 0.851}\textbf{7.412e+0} \\
		& 5     & 3.169e+1 & 3.169e+1 & 3.169e+1 & \cellcolor[rgb]{0.851, 0.851, 0.851}\textbf{3.169e+1} \\
		DTLZ4 & 8     & 2.558e+2 & 2.558e+2 & \cellcolor[rgb]{0.851, 0.851, 0.851}\textbf{2.558e+2} & 2.558e+2 \\
		& 10    & 1.024e+3 & 1.024e+3 & 1.024e+3 & \cellcolor[rgb]{0.851, 0.851, 0.851}\textbf{1.024e+3} \\
		& 15    & 3.276e+4 & 3.276e+4$^\dagger$ & 3.276e+4 & \cellcolor[rgb]{0.851, 0.851, 0.851}\textbf{3.276e+4} \\
		\hline
	\end{tabular}%
	\begin{tablenotes}
		\item[1] According to Wilcoxon's rank sum test, $^\dagger$ and $^\ddagger$ indicates whether the corresponding algorithm is significantly worse or better than MOEA/AD respectively.
	\end{tablenotes}
\end{table}

The comparison results between MOEA/AD and its three variants are presented in \pref{tab:DTLZ_v}. We find that our proposed MOEA/AD is still the best candidate where it obtains the best mean HV values on 16 out of 20 problem instances. In particular, its better HV values are with statistical significance on almost all DTLZ1 and DTLZ3 instances. Although MOEA/AD is outperformed by its variants on some instances, the differences to the best results are quite small. We have the following three assertions from the comparison results.
\begin{itemize}
	\item The stable matching procedure divides solutions of two populations into different pairs according to their working regions of the PF. It facilitates the mating selection process and help spread the search efforts along the while PF. In the meanwhile, the solution pair also provides some addition information on whether the paired solutions work on the similar regions.
	\item The collaboration between two populations help strengthen their complementary behaviors, i.e., one is diversity oriented and the other is convergence oriented.
	\item The three criteria together help to select a promising principal parent solution from a matching pair, which makes the reproduction more efficient.
\end{itemize}


\section{Conclusion}
\label{sec:conclusion}

In this paper, we have proposed MOEA/AD, a many-objective optimization algorithm based on adversarial decomposition. Specifically, it maintains two co-evolving populations simultaneously. Due to the use of different scalarizing functions, these two co-evolving populations have adversarial search directions which finally results in their complementary behaviors. In particular, one is convergence oriented and the other is diversity oriented. The collaboration between these two populations is implemented by a restricted mating selection scheme. At first, solutions from the two populations are stably matched into different one-one solution pairs according to their working regions. During the mating selection procedure, each matching pair can at most contribute one mating parent for offspring generation. By doing this, we can expect to avoid allocating redundant computational resources to the same region of the PF. By comparing the performance with nine state-of-the-art many-objective optimization algorithms on 130 problem instances, we have witnessed the effectiveness and competitiveness of MOEA/AD for solving many-objective optimization problems with various characteristics and PF's shapes. As a potential future direction, it is interesting to develop some adaptive methods that determine the scalarizing functions of different subproblems according to the PF's shape. It is also valuable to apply our proposed algorithm to other interesting application problems.

\bibliographystyle{IEEEtran}
\bibliography{IEEEabrv,MOEAD2P}

\end{document}